\documentclass[10pt,twocolumn,letterpaper]{article}

\usepackage{iccv}
\usepackage{times}
\usepackage{epsfig}
\usepackage{graphicx}
\usepackage{amsmath}
\usepackage{amssymb}
\usepackage{amsthm}
\usepackage{booktabs,tabularx}
\usepackage[format=plain,labelformat=simple,labelsep=period,font=small,skip=4pt,compatibility=false]{caption}
\usepackage[font=footnotesize,skip=2pt]{subcaption}
\usepackage[pagebackref=true,breaklinks=true,letterpaper=true,colorlinks,bookmarks=false]{hyperref}
\usepackage{verbatim}
\usepackage[inline]{enumitem}
\usepackage{mathtools}
\usepackage[ruled,linesnumbered]{algorithm2e}

\SetCommentSty{mycommfont}
\SetKwFunction{rgbtohsv}{rgb2hsv}
\SetKwFunction{medianfilter}{medianfilter}
\SetKwFunction{isolationforest}{isolationforest}
\usepackage[capitalize]{cleveref}
\crefname{table}{Tab.}{Tabs.}
\crefname{section}{Sec.}{Secs.}
\Crefname{table}{Table}{Tables}
\Crefname{section}{Section}{Sections}
\usepackage{etoolbox,siunitx}
\usepackage{nicefrac}
\newtheorem{theorem}{Theorem}[section]

\newtheorem{lemma}[theorem]{Lemma}
\robustify\bfseries
\robustify\itshape
\usepackage[keeplastbox]{flushend}
\usepackage{newtxtt}
\usepackage{appendix}

\newcommand{\myparagraph}[1]{\smallskip\noindent\textbf{#1}\hspace{0.25em}}

\usepackage[breaklinks=true,bookmarks=false]{hyperref}

\iccvfinalcopy 

\def\iccvPaperID{6285} 

\ificcvfinal\fi

\usepackage{fancyhdr}
\usepackage{setspace}

\begin{document}
\title{PixelPyramids: Exact Inference Models from Lossless Image Pyramids}

\author{
Shweta Mahajan$^1$ \qquad Stefan Roth$^{1,2}$ \\
 $\ ^1$Department of Computer Science, TU Darmstadt \qquad $\ ^2$ hessian.AI}


\maketitle
\thispagestyle{fancy}
\begin{abstract}
  Autoregressive models are a class of exact inference approaches with highly flexible functional forms, yielding state-of-the-art density estimates for natural images.
  Yet, the sequential ordering on the dimensions makes these models computationally expensive and limits their applicability to low-resolution imagery. 
  In this work, we propose \emph{Pixel\-Pyramids},\footnote{Code available at \url{https://github.com/visinf/pixelpyramids}} a block-autoregressive approach employing a lossless pyramid decomposition with scale-specific representations to encode the joint distribution of image pixels.
  Crucially, it affords a sparser dependency structure compared to fully autoregressive approaches. 
  Our PixelPyramids yield state-of-the-art results for density estimation on various image datasets, especially for high-resolution data.
  For CelebA-HQ $1024 \times 1024$, we observe that the density estimates  (in terms of bits/dim) are improved to $\sim\!\!44\,\%$  of the baseline despite sampling speeds superior even to easily parallelizable flow-based models.
\end{abstract}

\section{Introduction}
Deep generative models have enabled significant progress in capturing the probability density of highly structured, complex data, such as of natural images \cite{BrockDS19,GoodfellowPMXWOCB14,KingmaW13,RazaviOV19,OordKEKVG16} or raw audio \cite{DielemanOS18,HsuZWZWWCJCSNP19,OordLBSVKDLCSCG18}.
These models find application in a wide range of tasks including image compression \cite{CaoWK20,TschannenAL18}, denoising \cite{BalleLS15}, inpainting \cite{TheisB15,OordKK16}, and super-resolution \cite{BrunaSL15,LugmayrDGT20}. 
Popular generative models for images include Generative Adversarial Networks (GANs) \cite{GoodfellowPMXWOCB14}, which transform a noise distribution to the desired distribution, and Variational Autoencoders (VAEs) \cite{KingmaW13}, where the data is modeled in a low-dimensional latent space by maximizing a lower bound on the log-likelihood of the data \cite{RezendeMW14}.
However, GANs are not designed to provide exact density estimates and VAEs only approximate the underlying true distribution with intractable likelihoods, posing challenges in both training and inference.
\begin{figure}[t]
\centering
	\begin{subfigure}[b]{0.75\columnwidth}
        \centering
        \includegraphics[width=\linewidth]{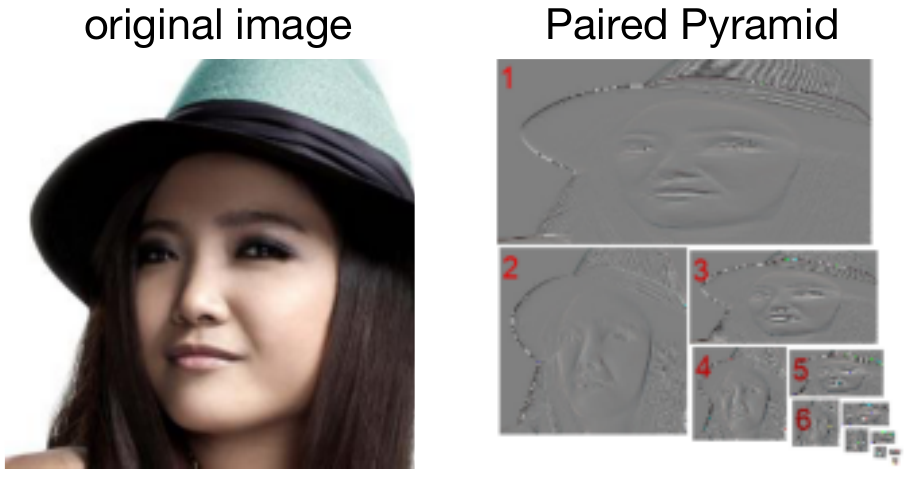}
    \end{subfigure}\\[-0.45em]
    \begin{subfigure}[b]{0.7\columnwidth}
        \centering
        \includegraphics[width=\linewidth]{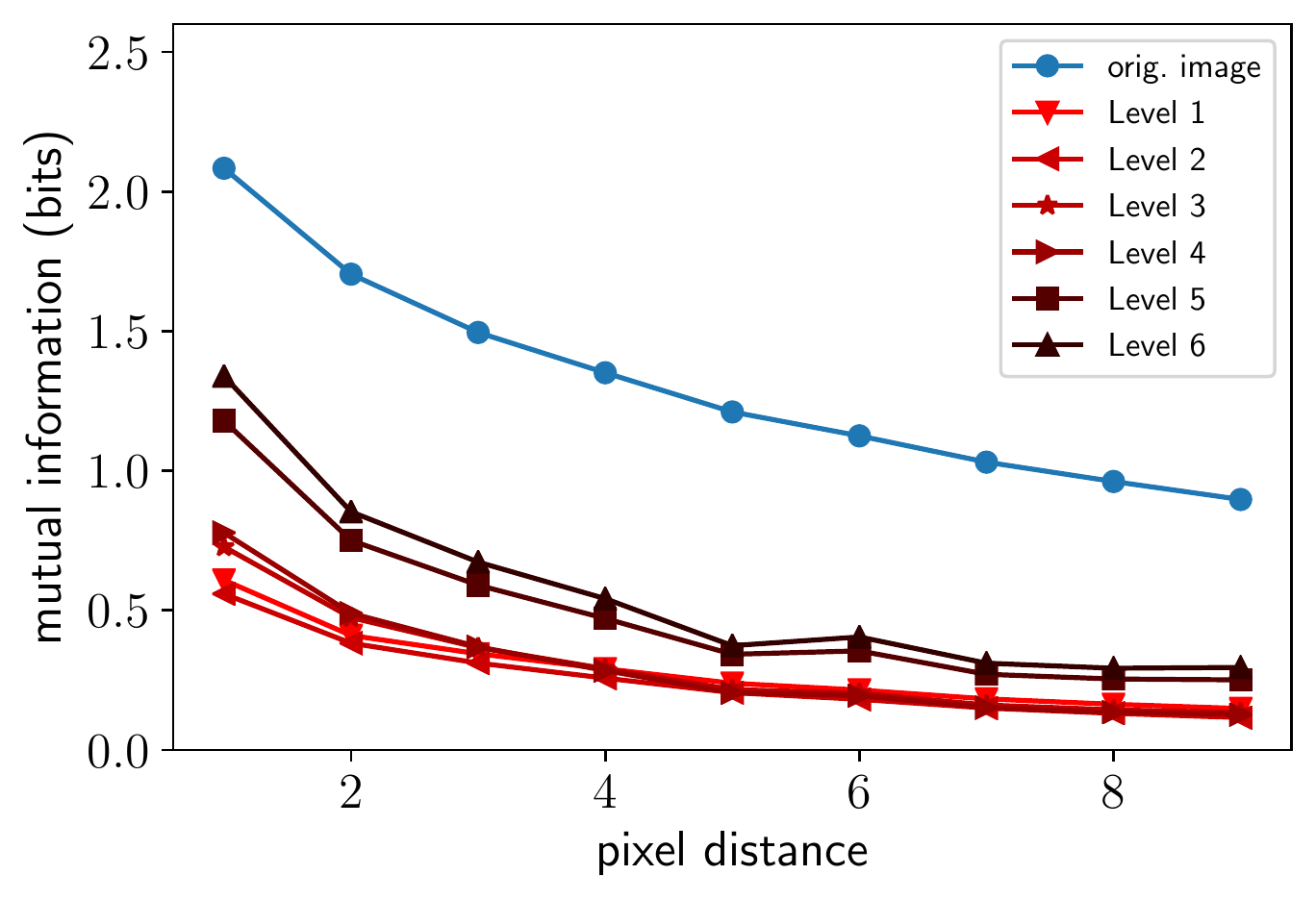}
    \end{subfigure}%
\caption{\emph{Motivation for PixelPyramids.}
An image from CelebA \cite{KarrasALL18} \emph{(top left)} is decomposed into a Paired Pyramid \cite{10.1117/12.970109} \emph{(top right)} through an invertible, lossless mapping.
Considering the mutual information between pixels as a function of pixel distance on the CelebA dataset \emph{(bottom)},
there is significant spatial dependence in the original image \emph{(blue, circles)}, even between distant pixels.
The fine components of the Paired Pyramid at different levels \emph{(other lines)} have significantly more localized dependencies, decreasing further as the resolution increases.
Our PixelPyramids exploit this for efficient block-autoregressive generative image modeling.}
\label{fig:mi_stats}
\vspace{-0.5 em}
\end{figure}

In contrast, autoregressive models \cite{domke2008killed,MenickK19,OordKEKVG16,OordKK16} and normalizing flows \cite{BhattacharyyaMF20,DinhKB14,DinhSB17,HoCSDA19,KingmaD18} are exact inference approaches, which estimate the exact likelihood of the data.
Autoregressive models factorize the joint target distribution into a product of conditional distributions with a certain ordering over the dimensions, thereby encoding complex dependencies in the data distribution for effective density estimation.
However, the sequential dependency structure makes efficient parallelization difficult.
Normalizing flows, on the other hand, map the input data to a known base distribution through a series of invertible transformations, providing efficient sampling.
However, the invertibility constraint limits their expressiveness and their density estimation performance lags behind that of autoregressive models.

Recent work \cite{MaKZH19,ReedOKCWCBF17,YuDB20} aims to improve the computationally expensive (autoregressive) or constrained (normalizing flows) exact inference models through multi-scale feature representations of images, such as that from Wavelet or pyramid representations.
The key idea is that instead of directly modeling the complex image distribution, it is decomposed into a series of simpler pixel representations, which can be encoded with comparatively less complex functional forms of exact inference models.
However, these approaches either rely on a specific design choice of pixel orderings to reduce autoregressive connections or contain higher quantization levels, which makes encoding difficult. 

In this work, we propose \emph{PixelPyramids} -- an expressive and computationally efficient block-autoregressive model for the discrete joint distribution of pixels in images.
We make the following contributions: 
\emph{(i)} Our PixelPyramids leverage ideas from lossless image coding, specifically a low-entropy multi-scale representation based on so-called Paired Pyramids \cite{10.1117/12.970109} (\cf \cref{fig:mi_stats}, \emph{top}), to encode images in a coarse-to-fine manner, where the conditional generative model at each pyramid scale is conditioned on the coarse(r) image from the previous level.
The subsampled images and the scale-specific components are encoded in the same number of pixels and at the same quantization level as the original image, resulting in an efficient exact inference approach for density estimation.\footnote{We use the common term `density estimation' even if we, strictly speaking, estimate a discrete distribution rather than a continuous density.}
\emph{(ii)} The conditional generative model for the fine component at each scale exploits a U-Net architecture \cite{RonnebergerFB15} to encode global context from the coarser scales.
A Convolutional LSTM \cite{ShiGL0YWW17} is used to capture the spatial pixel dependencies within each fine component.
By conditioning on the coarser scales, the spatial dependency structure of the fine component at each scale is more localized (\cf~\cref{fig:mi_stats}, \emph{bottom}), especially as we go to higher resolutions in the pyramid.
Thus, long-range dependencies at each scale can be modeled with fewer autoregressive steps in a computationally efficient setup.

Furthermore, \emph{(iii)} a key advantage of PixelPyramids is that the fine components at different scales are conditionally independent of each other and can, therefore, be trained in parallel.
This makes the model applicable for density estimation and synthesis of high-resolution images.
PixelPyramids require $\mathcal{O}(\log N)$ sampling steps for an $N \times N$ image, which is significantly more efficient than the $\mathcal{O}(N^2)$ sequential steps of fully autoregressive approaches.
Finally, \emph{(iv)} we show that our PixelPyramids yield state-of-the-art density estimates and high-quality image synthesis on standard image datasets including CelebA-HQ \cite{KarrasALL18}, LSUN \cite{YuZSSX15}, and ImageNet \cite{Russakovsky:2015:ILS}, as well as on the high-resolution $1024 \times 1024$ CelebA-HQ dataset at a much lower computational cost than previous exact inference models.

\section{Related Work}
Exact inference models explicitly maximize the data log-likelihood, making them attractive for estimating multimodal densities.
Recent work on deep generative models applicable to both density estimation and generation has focused on autoregressive methods and normalizing flows.

\myparagraph{Normalizing flows} map a simple base distribution to the complex data distribution through a series of invertible transformations with tractable Jacobian determinants~\cite{ArdizzoneLKRK19,DinhKB14,kingma2016improved}.
Kingma \etal~\cite{KingmaD18} extend the flow-based RealNVP model \cite{DinhSB17} with invertible $1 \times 1$ convolutions with applications to high-resolution imagery.
Various work \cite{HoCSDA19,HoogeboomBW19,MaKZH19} improves flow-based models with architectural modifications to the non-linearities in the invertible transformations to increase their modeling capacity or provide for efficient training.
While easy to parallelize and thus efficient, the representational power of flow-based models reflected in density estimates lags behind that of autoregressive models.

\myparagraph{Autoregressive models} 
factorize the image distribution into a product of conditionals over the pixels, exploiting the chain rule for exact inference~\cite{domke2008killed,Graves13,HochreiterS97,ParmarVUKSKT18}.
Deep neural generative autoregressive models, \eg~the PixelCNN and PixelRNN of van den Oord \etal\ \cite{OordKEKVG16,OordKK16}, parameterize the conditionals with convolutional or recurrent neural networks, achieving state-of-the-art density estimates on natural images.
Kalchbrenner \etal~\cite{KalchbrennerESN18} apply autoregressive models to raw audio. 
Subsequent work of Chen \etal\ and Salimans \etal\ \cite{ChenMRA18,SalimansK0K17} introduces architectural enhancements to PixelCNN such as self-attention layers \cite{ChenMRA18} and a discretized logistic mixture model \cite{SalimansK0K17}.
One of the main limitations of these approaches is that generation proceeds through ancestral sampling, making them slow and difficult to parallelize.
Song \etal\ \cite{song21a} and Wiggers \etal\ \cite{WiggersH20} marginally improve the sampling speed by parallelizing the Jacobi updates with Gauss-Seidel and fixed-point iterations, respectively, subject to the availability of specific hardware resources.
Noting the current limitations in modeling sequential dependencies in a computationally efficient manner, we introduce a multi-scale coarse-to-fine structure with localized dependency structures to model the discrete joint distribution of image pixels.

\begin{figure*}[t!]
\begin{subfigure}[t]{0.45\linewidth}
\centering
   \includegraphics[width=\linewidth]{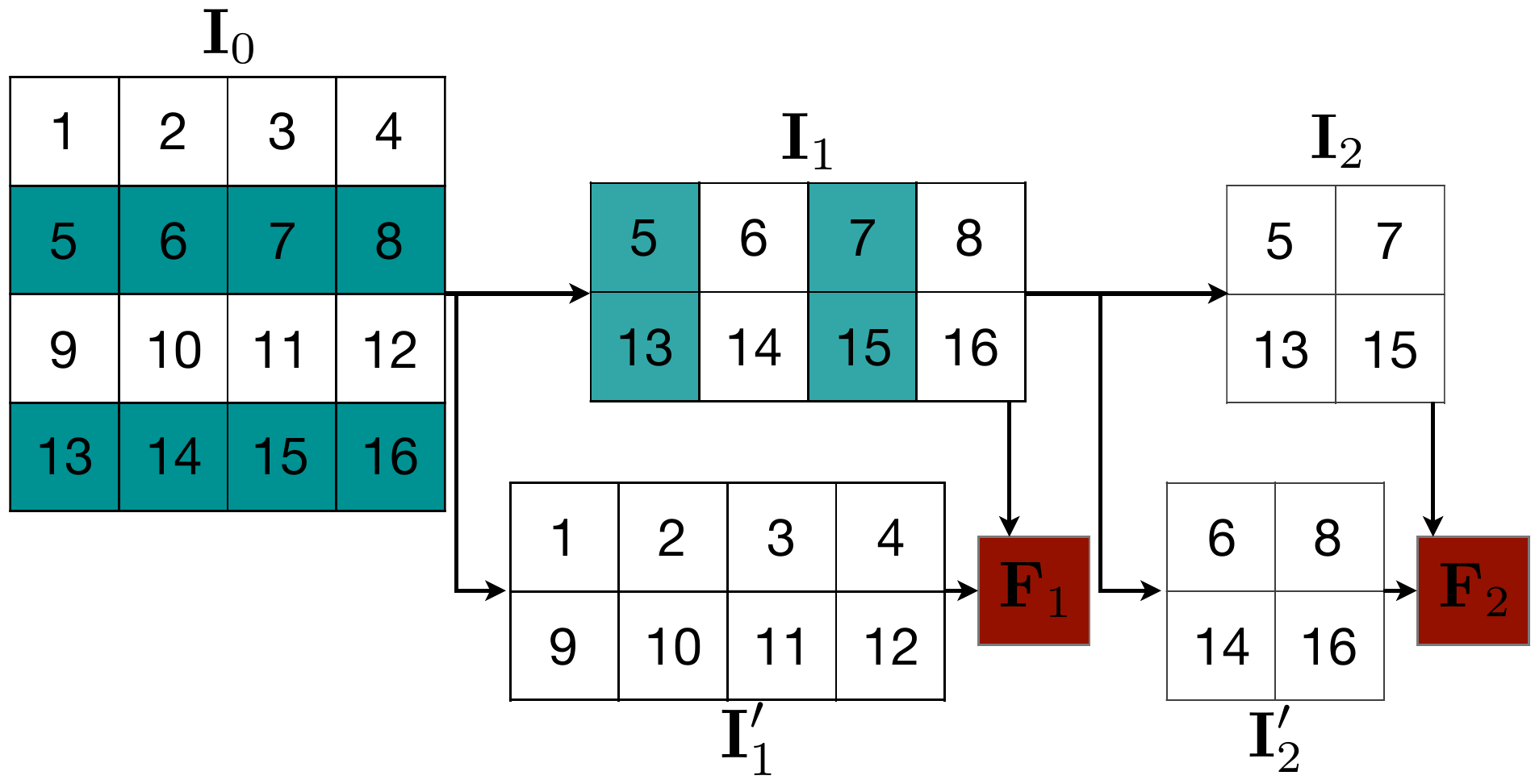}
   \caption{Two levels of a Paired Pyramid decomposition of an image}
   \label{fig:pyramid_iter}
\end{subfigure}%
\hfill
\begin{subfigure}[t]{0.5\linewidth}
\centering
  \includegraphics[width=\linewidth]{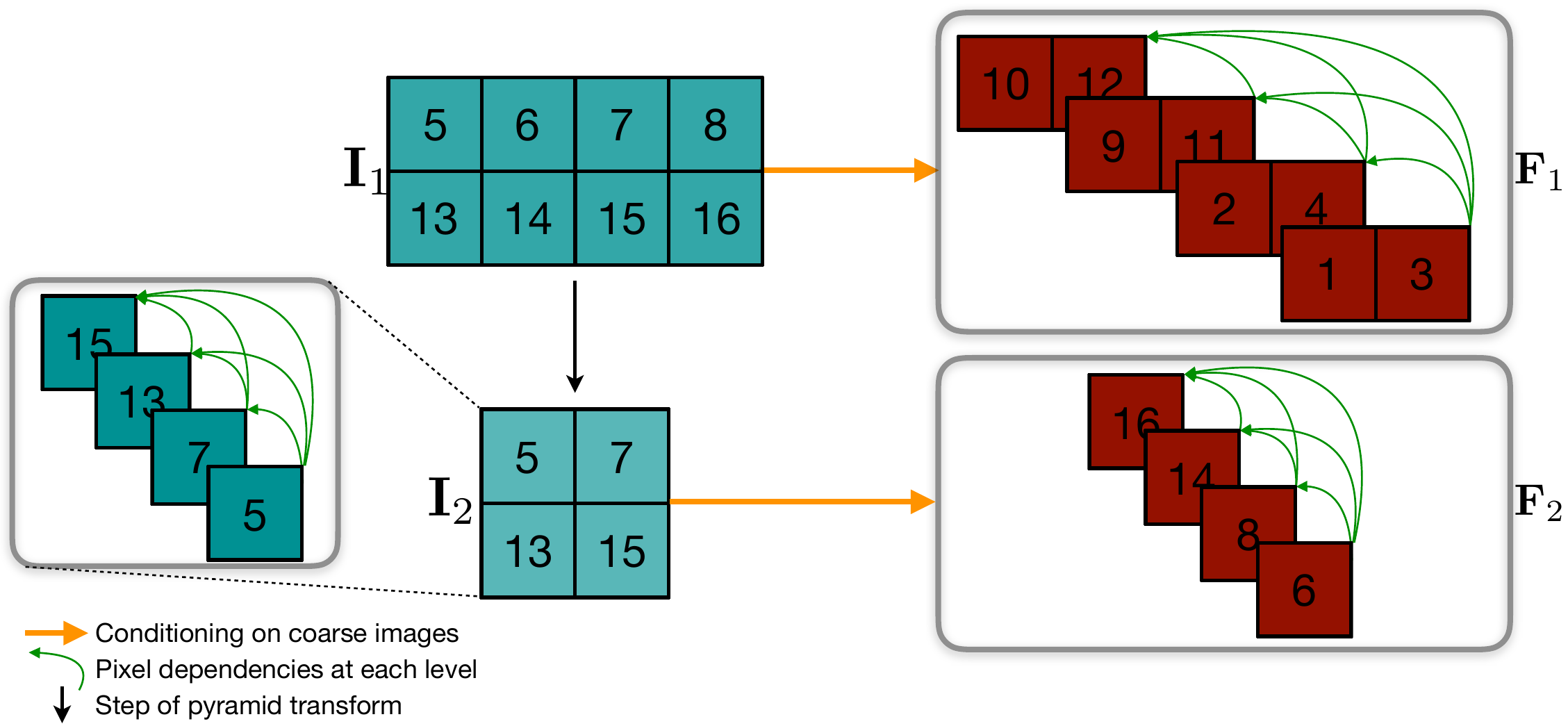}
  \caption{Spatial dependencies at different levels}
  \label{fig:pyr_dep}
\end{subfigure}
\caption{\emph{Paired Pyramids and the imposed dependency structure.}
\emph{(left)} Subsampling procedure in a Paired Pyramid (numbers denote pixel indices). 
In the first iteration, $\mathbf{I}_0$ is subsampled along the rows of the image, yielding $\mathbf{I}_1$ and $\mathbf{I}'_1$. 
The fine component $\mathbf{F}_1$ is the modulo difference of $\mathbf{I}'_1$ and $\mathbf{I}_1$, see \cref{eq:mod_diff}. 
In the second iteration, $\mathbf{I}_1$ is subsampled along the columns, yielding $\mathbf{I}_2$ and $\mathbf{I}'_2$.
\emph{(right)} The spatial dependencies are modeled densely at the coarsest level (fully autoregressive) and more localized in pixel groups at higher, finer levels.}
\label{fig:pyr}
\end{figure*}

\myparagraph{Multi-scale representations} have been leveraged in GANs and VAEs for encoding the multimodality of the data distribution. 
Denton \etal~\cite{DentonCSF15} use residuals from Laplacian pyramids \cite{BurtA83} for multi-stage generation in GANs.
Dorta \etal~\cite{DortaVACPS17} utilize detail coefficients from Laplacian pyramids in the latent space of VAEs for image manipulation.
Razavi \etal~\cite{RazaviOV19} extend VQ-VAEs with hierachical low-level and high-level details in the latent space. In this work, we consider models that provide for exact inference in contrast.

Multiscale-PixelCNNs \cite{ReedOKCWCBF17} aim to improve the sampling speed of autoregressive models with strong independence assumptions on the conditionals in the subsampled images.
Menick \etal~\cite{MenickK19} synthesize by sequential conditioning on sub-images and employ an autoregressive structure not only along the spatial dimensions but also along image pixel values.
Both methods rely on a design choice of the pixel ordering for generating multiple levels of the pyramid, which does not take into account image statistics such as harnessing scale-specific information.
Kolesnikov \etal~\cite{KolesnikovL17} introduce conditionals with grayscale image representations to capture global context.
The conditionals in these multi-scale autoregressive models are applied directly in the pixel space and, therefore, have redundancies in the dependency structure at different scales.
Recent work of Bhattacharyya \etal\ \cite{BhSFB20} and Yu \etal\ \cite{YuDB20} increases the modeling flexibility of flow-based models with multi-scale coarse-to-fine scale-space representations based on Haar Wavelets. 
In our work, to improve the modeling flexibility and computational efficiency of autoregressive models, we leverage multi-scale lossless decompositions, which provide localized scale-specific representations and can be encoded with a limited number of autoregressive connections.

\section{Block-Autoregressive PixelPyramids}
To learn the complex joint distribution of image data with exact inference and generation in a computationally efficient framework, we formulate a low-entropy multi-scale block-autoregressive approach termed \emph{PixelPyramids}. 
Our method leverages Paired Pyramids \cite{10.1117/12.970109} from the literature on lossless image coding to decompose the image into low-entropy components with a local dependency structure, which makes it easier to perform density estimation.

\subsection{Paired Pyramids}
Proposed by Torbey and Meadows \cite{10.1117/12.970109}, Paired Pyramids are an invertible, lossless transformation of the image $\mathbf{I}_0$ of size ${N_0 \times N_0 \times C}$ into $L$ levels of fine and coarse components.
One of the spatial dimensions is reduced by a factor of two at each level of the image pyramid by alternatingly subsampling along the rows and columns of the image.
To construct the \emph{coarse component} $\mathbf{I}_i$ at level $i \in \{1,\ldots,L\}$, one row (column) from each of the non-overlapping pairs of adjacent rows (columns) of the previous level $\mathbf{I}_{i-1}$ is retained without any transformation to form a subsampled image (see \cref{fig:pyramid_iter}). This reduces the spatial dimension by a factor of two, \ie to size $\nicefrac{N_{i-1}}{2} \times N_{i-1} \times C$ when subsampled along the rows (analogously for subsampling along the columns). 
The image $\mathbf{I}'_{i}$ consisting of the remaining rows (columns) from each pair, also with size $\nicefrac{N_{i-1}}{2} \times N_{i-1} \times C$, is used to encode the \emph{fine component} $\mathbf{F}_i$ through the \emph{modulo difference} between the adjacent pairs of rows (columns) as
\begin{align}
\mathbf{F}_i = (\mathbf{I}'_{i} - \mathbf{I}_i) \bmod 2^{b}.
\label{eq:fine_comp}
\end{align}
Here, $b$ is the number of bits for encoding the image $\mathbf{I}_0$, \ie $\mathbf{I}_0\in\{0,\ldots,2^b-1\}^{N_0\times N_0\times C}$, and its components $\mathbf{I}_i$, $\mathbf{I}'_i$, or $\mathbf{F}_i$.
The modulo $K$ difference, with $x, y \in \{0,\ldots, K-1\}$, can be computed as 
\begin{align}
(x - y) \bmod K  =
\begin{cases}
(x - y) & \text{if $ x \geq y$} \\
K-(y - x)& \text{otherwise}.
\end{cases}
\label{eq:mod_diff}
\end{align}
Thus, the modulo difference of the two subsampled images lies in the same range as the subsampled or the original image.
Furthermore, the subsampled component $\mathbf{I}'_i$ can be fully reconstructed given the fine component $\mathbf{F}_{i}$ and the coarse component $\mathbf{I}_i$ using the inverse transformation
\begin{align}
\mathbf{I'}_i = (\mathbf{F}_{i} + \mathbf{I}_i) \bmod 2^{b}.
\label{eq:inv_mod_diff}
\end{align}

\begin{figure*}[t]
    \centering
   \includegraphics[width=0.75\textwidth]{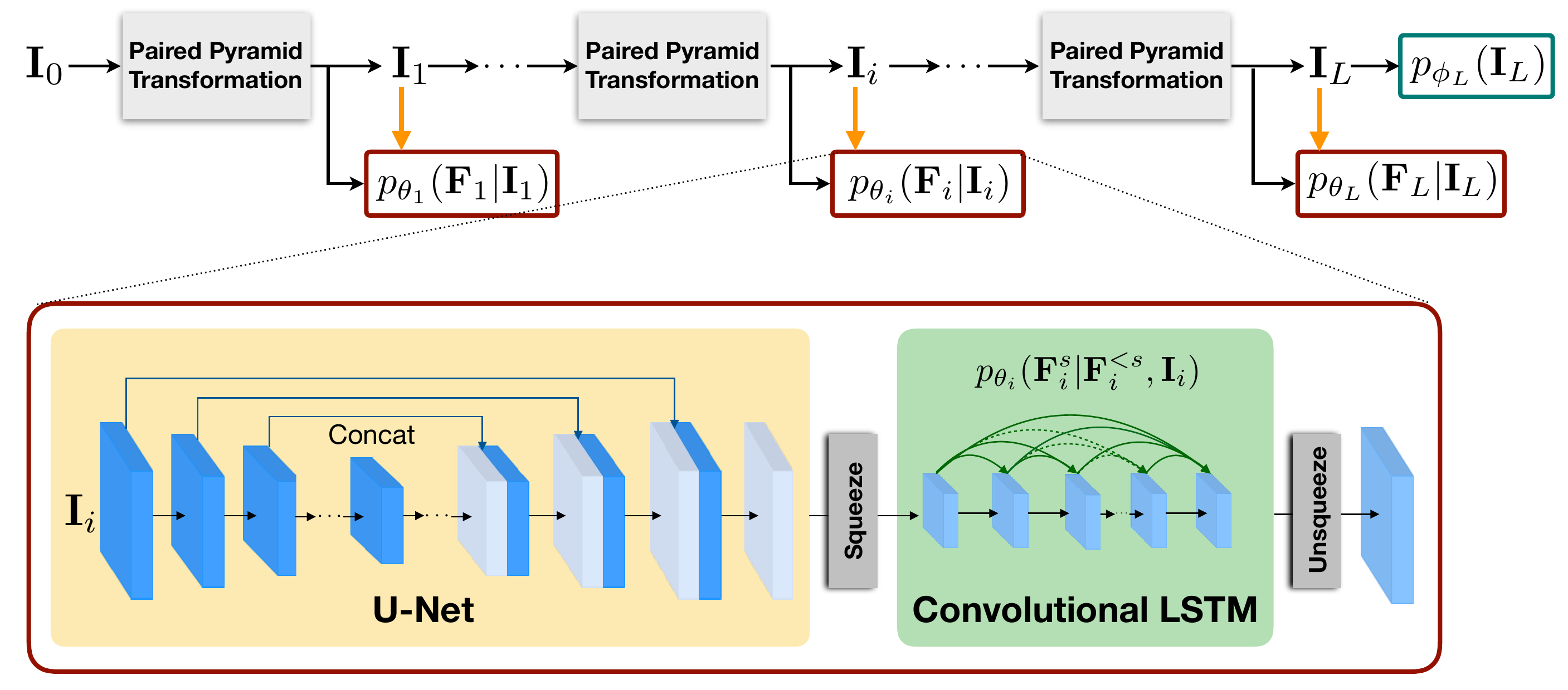}
    \caption{\emph{PixelPyramids -- Multi-scale block-autoregressive generative model based on Paired Pyramids} \cite{10.1117/12.970109}. The proposed PixelPyramids rely on a U-Net \cite{RonnebergerFB15} for capturing coarse-scale context, conditioned on which Conv-LSTMs \cite{ShiGL0YWW17} emit the fine components of the pyramid.}
    \label{fig:archi}
    \vspace{-0.5em}
\end{figure*}

\begin{figure}[t]
\begin{subfigure}[t]{0.45\columnwidth}
\centering
   \includegraphics[width=\linewidth]{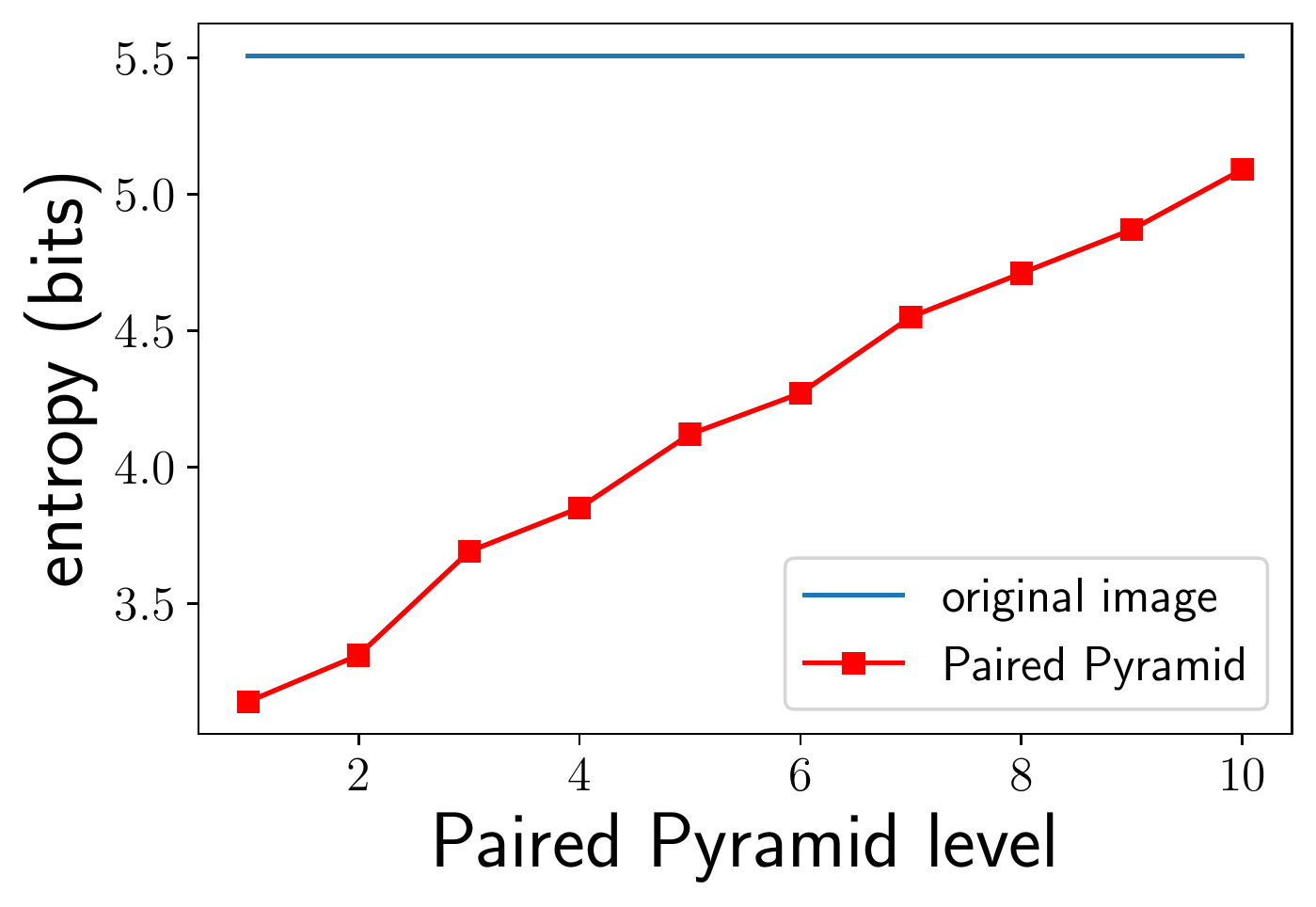}
\end{subfigure}%
\hfill
\begin{subfigure}[t]{0.45\columnwidth}
\centering
  \includegraphics[width=\linewidth]{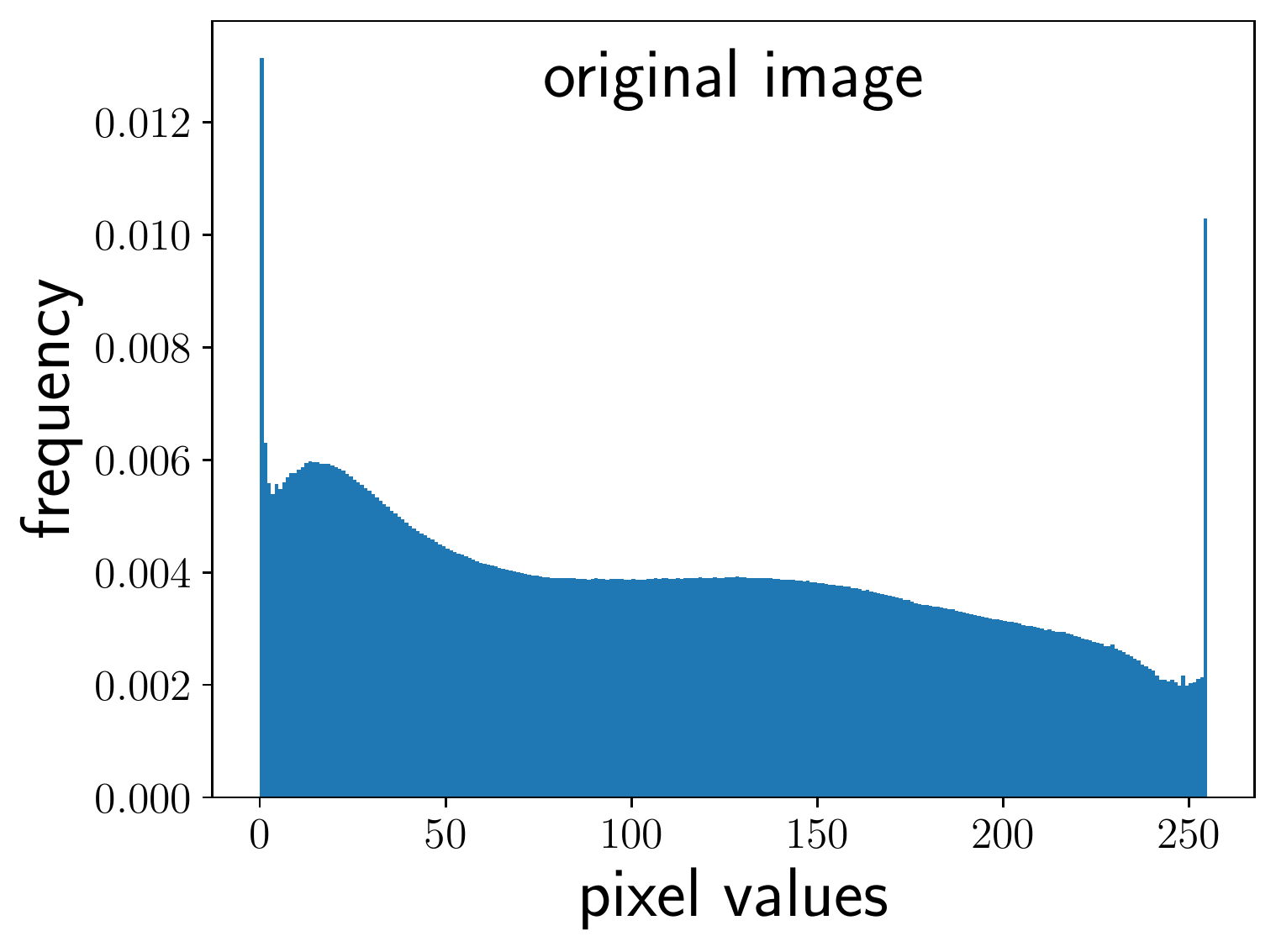}
\end{subfigure}
\begin{subfigure}[t]{0.45\columnwidth}
\centering
  \includegraphics[width=\linewidth]{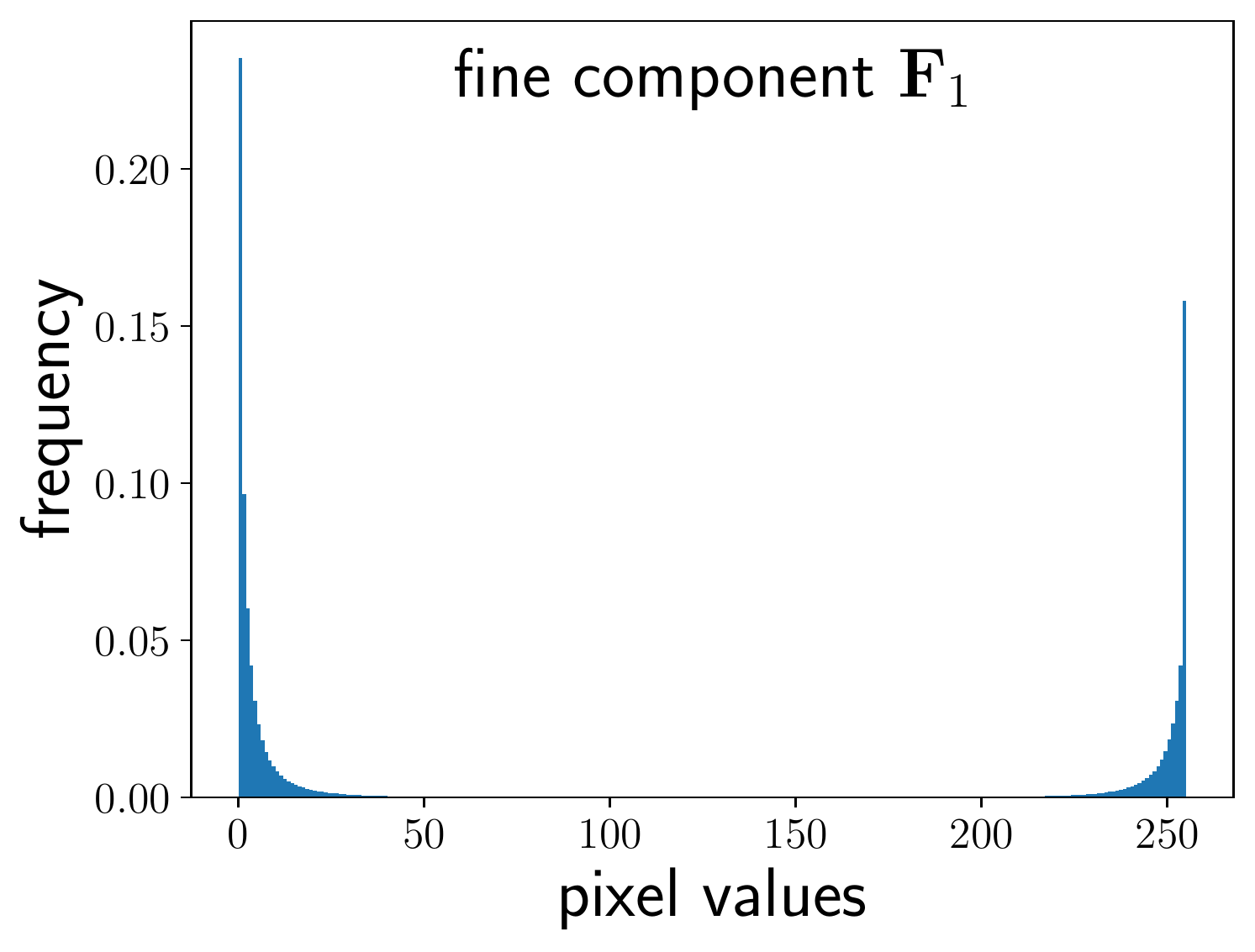}
\end{subfigure} 
\hfill
\begin{subfigure}[t]{0.45\columnwidth}
\centering
  \includegraphics[width=\linewidth]{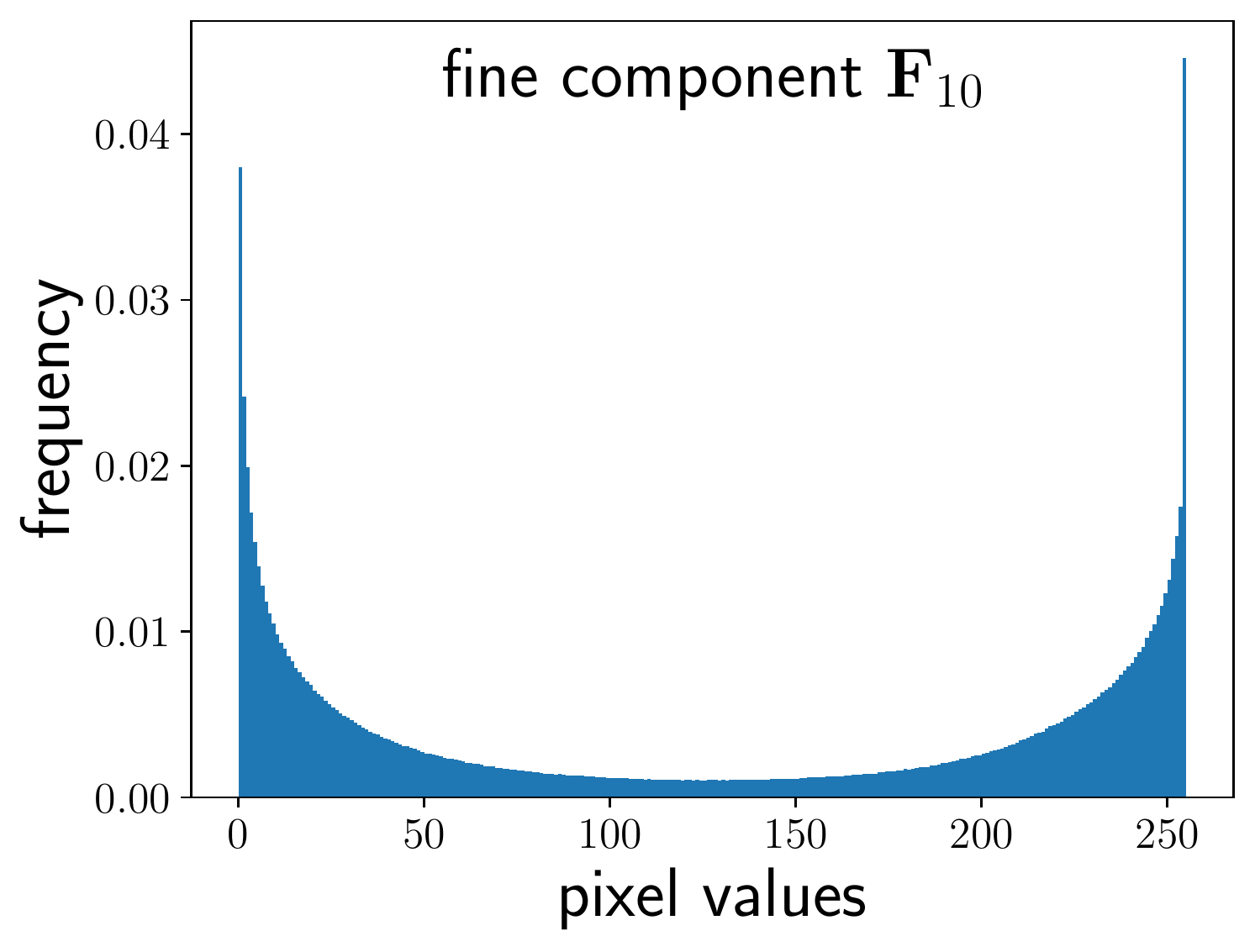}
\end{subfigure}
\caption{\emph{Low-entropy decomposition.} Entropy values at different pyramid levels \emph{(top left)}. Empirical marginal distributions on CelebA-HQ for the original pixel values \emph{(top right)}, the pixel values of the fine component $\mathbf{F}_{1}$ \emph{(bottom left)} and and the fine component $\mathbf{F}_{10}$ \emph{(bottom right)}.
Note the significantly reduced entropy from 5.51 bits for the original pixel values to 3.14 bits for $\mathbf{F}_1$.}
\label{fig:hist_marginal}
\vspace{-0.5em}
\end{figure}

\myparagraph{Representational properties.}
Paired Pyramids have various favorable properties:
\emph{(i)} Using the modulo operation, the fine components can be represented at the \emph{same quantization level} as the original image (no additional bit needed to represent the sign \cite{10.1117/12.970109}).
\emph{(ii)} Moreover, the resulting pyramid representation has the \emph{same number of pixels} as the original image (see \cref{fig:mi_stats}, \emph{top}). 
\emph{(iii)} This makes the modulo difference-based pyramid transform an \emph{invertible, lossless image coding scheme} \cite{10.1117/12.970109}.
Note that this is unlike the standard form of a Laplacian pyramid \cite{BurtA83}, which is overcomplete and the detail coefficients are difficult to represent in the same range and quantization level as the original image without loss of information.
Here, we leverage the local, scale-specific representations of Paired Pyramids to develop a computationally efficient deep generative model with tractable likelihoods for complex image distributions. 


\myparagraph{Statistical properties.}
To see why Paired Pyramids are well suited also for generative modeling, we consider their statistical properties.
We observe the following:
\emph{(i)} The fine components $\mathbf{F}_i$ at each level represent the scale-specific image structure and, as shown in \cref{fig:mi_stats} \emph{(bottom)}, have \emph{significantly less pixel-to-pixel correlations} than the original image.
\emph{(ii)} Furthermore, as shown in \cref{fig:hist_marginal}, its pixel values are highly concentrated around 0 and $2^{b}-1$, \ie \emph{have low entropy} compared to the much more uniform marginal pixel distribution of the original images.
Both the low entropy and the low spatial correlation make the fine components substantially easier to model than the original image.
\emph{(iii)} Importantly, entropy and spatial correlation decrease further at lower (high res.) pyramid levels, which means that \emph{the computationally most costly levels are the easiest to model}.

\subsection{Pixel Pyramid Network}
To define PixelPyramids (\cref{fig:archi}), we first construct a Paired Pyramid representation of the original image $\mathbf{I}_0$, consisting of $L$ levels with coarse components \{$\mathbf{I}_1,\ldots, \mathbf{I}_L$\} and fine components \{$\mathbf{F}_1,\ldots, \mathbf{F}_L$\}. 
Equipped with this lossless (one-to-one) decomposition, the probability distribution $p_{\theta,\phi_L}(\mathbf{I}_0)$ of images, parameterized by $\theta=\{\theta_1,\ldots,\theta_L\}$ and $\phi_L$, autoregressively factorizes \textit{w.l.o.g.}~as
\begin{align}
p_{\theta, \phi_L}(\mathbf{I}_0) = p_{\phi_L}(\mathbf{I}_L) \prod \limits_{i=1}^{L} p_{\theta_i}(\mathbf{F}_i \mid \mathbf{I}_i).
\label{eq:pp_net}
\end{align}
Thus, the image distribution can be obtained as the product of the distribution of the coarsest component $\mathbf{I}_L$ and the conditional distributions of the fine components $\mathbf{F}_i$ conditioned on the coarse component $\mathbf{I}_i$ at each level.
The main advantage of this decomposition is three-fold:
\emph{(i)} The fine components have \emph{low marginal entropy}, thus are easier to model using parameterized conditional deep generative models;
\emph{(ii)} the fine components also have \emph{less spatial correlations} and represent scale-specific local information; 
\emph{(iii)} the spatial correlations of the fine components are \emph{less at the} (computationally challenging) \emph{higher resolution levels} than at the (computationally lighter) lower resolution levels. 
Therefore, the conditional distribution $p_{\theta_i}(\mathbf{F}_i \mid \mathbf{I}_{i})$ becomes easier to model with increasing scales $i$ using computationally efficient models with local pixel dependencies.
The factorization in \cref{eq:pp_net} has yet another advantage:
The conditional generative model of the fine component at every scale is independent of those at the other scales.
Thus they can be trained in parallel.
Taken together, this overcomes the limitations of models like PixelCNN \cite{OordKEKVG16} or SPN \cite{MenickK19}, which rely on a fully autoregressive structure to capture the underlying multimodality and complex pixel dependencies.

\myparagraph{Per-level conditionally generative models.}
In addition to conditioning on the coarse image $\mathbf{I}_i$, we introduce a partial autoregressive structure to better encode the dependencies within the fine component $\mathbf{F}_i$.
To that end, $\mathbf{F}_i$ is further decomposed into subsampled images (see \cref{sec:network-archi} for details), where we enforce sequential dependencies across the pixels of different subsampled images while the pixels within each subsampled image are conditionally independent of each other.
This is illustrated in \cref{fig:pyr_dep} \emph{(top)} for the case of four such subsampled images.
Let $S_i$ be the number of subsampled images in the fine component at a certain level $i$ and $\mathbf{F}_i^{s}, s \in \{1,\ldots,S_i\}$ denote a subsampled image. 
Then the conditional distribution $p_{\theta_{i}}(\mathbf{F}_i \mid \mathbf{I}_i)$ decomposes into a product of conditionals using the product rule as
\begin{align}
    p_{\theta_{i}}(\mathbf{F}_i \mid \mathbf{I}_i) =  p_{\theta_{i}}(\mathbf{F}_i^{1} \mid \mathbf{I}_i) \prod_{s=2}^{S_i} p_{\theta_{i}}(\mathbf{F}_i^{s} \mid \mathbf{F}_i^{<s}, \mathbf{I}_i).
    \label{eq:per-level-conditionals}
\end{align}
The benefit of the Paired Pyramid representation is that we do not need many such autoregressive sampling sub-images, since the fine components are spatially decorrelated, especially those at higher resolutions (see above).

\myparagraph{Modeling the coarsest level.}
The image at the coarsest level $L$ has size $N_0/{2^{\lfloor (L+1)/2 \rfloor}} \times N_0/{2^{\lfloor L/2 \rfloor}} \times C$ and represents the global structure at a much lower resolution than the original image.
Assuming that we have sufficiently many levels, the coarsest level is small enough to adopt a fully autoregressive model where, similar to PixelCNN, pixels $\mathbf{I}_L^{s}$ are generated one-by-one using the chain rule, conditioning on both pixels in the rows above it and to the left within the same row, $\mathbf{I}_L^{<s}$ (\cref{fig:pyr_dep}, \emph{bottom}).
We thus have
\begin{align}
 p_{\theta_{L}}(\mathbf{I}_L) = p_{\theta_{L}}(\mathbf{I}_L^{1}) \prod_{s=2}^{S_L} p_{\theta_{L}}(\mathbf{I}_L^{s} \mid \mathbf{I}_L^{<s}).
\end{align}

\myparagraph{Per-pixel conditional distributions.}
The pixel values for the fine components at each level as well as the image at the coarsest level are modeled as discrete distributions.
Note that due to the properties of the Paired Pyramid, all components have the same range and quantization as the original image. 
Similar to PixelCNN++ \cite{SalimansK0K17}, we use a continuous mixture of $M$ logistic distributions to estimate the conditional distribution of a pixel value $x\in\{0,\ldots,2^b-1\}$ of the fine or coarse component as
\begin{align}
\begin{split}
    p(x \mid \boldsymbol{\pi},\boldsymbol{\mu},\boldsymbol{\omega})
    = \sum_{i=1}^M \pi_i\Big[&\sigma\big((x+0.5-\mu_i)/\omega_i\big)\\[-1em]
    &-\sigma\big((x-0.5-\mu_i)/\omega_i\big)\Big].
\end{split}
\label{eq:per-pixel-conditionals}
\end{align}
Here, $\sigma(\cdot)$ is the logistic function, $\mu_i$ and $\omega_i$ are the conditional mean and standard deviation of each mixture component with weight $\pi_i$.
The mean of the green channel is taken to linearly depend on the value of the red sub-pixel and, analogously, the mean of the blue channel on the red and green sub-pixel values.
For instantiating a pixel value from the continuous logistic mixture, the samples are rounded to the nearest discrete value with the same quantization as the fine component.
We refer to \cite{SalimansK0K17} for more details. 

\subsection{Network architecture}
\label{sec:network-archi}
To yield the conditional distributions $p_{\theta_i}(\mathbf{F}_i \mid \mathbf{I}_i)$, we first leverage a U-Net architecture \cite{RonnebergerFB15} for capturing the conditional information from $\mathbf{I}_i$ (\cref{fig:archi}).
Specifically, its low-resolution feature maps in the encoder provide global context from the conditioning pixels for encoding scale-specific image details.
Combining this through skip connections with the upsampled features of the decoder adds local context at higher resolutions (see \cref{sec:experiments-archi} for details).
The feature map obtained from the final U-Net layer, with the same spatial dimensions as that of the fine component, is then used as input for modeling $\mathbf{F}_i$ following \cref{eq:per-level-conditionals}.

To that end, we decompose the image/feature map into subsampled images.
Specifically, we apply a number of squeeze operations \cite{DinhKB14} to trade off spatial dimensions for channel dimensions.
The squeeze operation stacks non-overlapping $2\times 2 \times 1$ local neighborhoods along the feature channels by reshaping them into blocks of size $1 \times 1 \times 4$.
Applying this to non-overlapping neighborhoods from an $N \times N \times C$ image yields a representation of size $\nicefrac{N}{2} \times \nicefrac{N}{2} \times 4\cdot C$, thus containing four subsampled images each of size $\nicefrac{N}{2} \times \nicefrac{N}{2} \times C$.
This can be repeated and allows for a partial autoregressive structure, where different subsampled images are generated sequentially and the pixels within the subsampled image can be generated in parallel.

The long-range sequential dependencies between the subsampled images are modeled through the internal state of the convolutional LSTMs \cite{ShiGL0YWW17}.
Additionally, spatial dependencies within the subsampled images are captured by stacking multiple Conv-LSTM layers, providing a wide receptive field (see \cref{sec:experiments-archi} for details).
Following the Conv-LSTMs, unsqueeze operations bring the spatial resolution back to the size of the fine component, yielding the per-pixel parameters of the mixture distribution from \cref{eq:per-pixel-conditionals}.

\myparagraph{Sampling.}
Fully autoregressive approaches require $\mathcal{O}(N_0^2)$ sampling steps.
In contrast, our PixelPyramids -- given sufficient parallel resources to sample a channel in parallel -- require only a logarithmic number of steps:
\begin{lemma}
Let the sampled image $\mathbf{I}$ be of resolution $\left[N_0,N_0,C\right]$, then the worst-case number of steps $T$ (length of the critical path) required is $\mathcal{O}(\log N_0)$.
\end{lemma}

\noindent\emph{Proof.}
See supplemental material.


\begin{figure}[t]
    \centering
    \includegraphics[width=0.95\columnwidth]{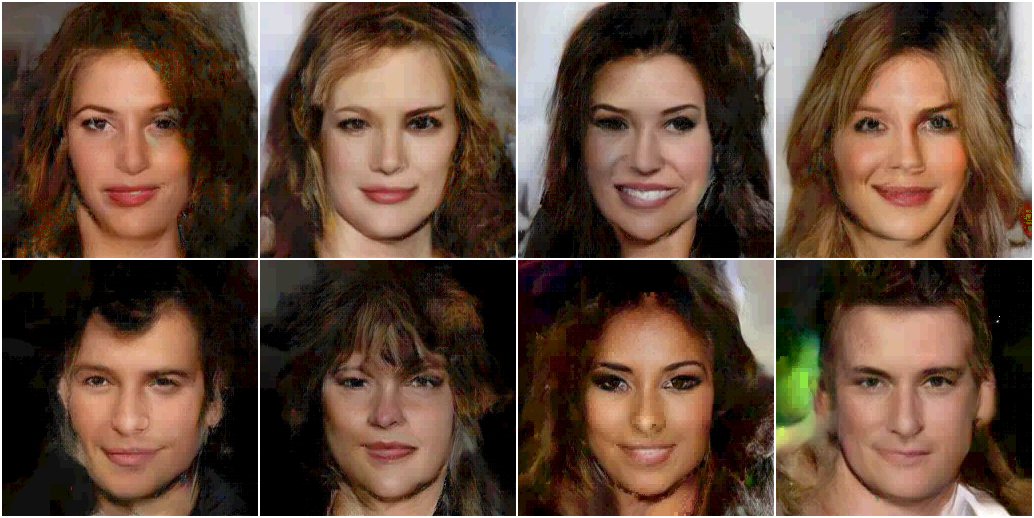}
    \caption{Random samples from 5-bit CelebA-HQ ($256 \times 256$).}
    \label{fig:celeba256_samp}
    \vspace{-0.5em}
\end{figure}

\begin{table}[t]
  \centering
  \scriptsize
  \begin{tabularx}{\columnwidth}{@{}X@{\hskip 1.5em}S[table-format=1.2]S[table-format=1.2]S[table-format=1.2]S[table-format=1.2]@{}}
	\toprule
	  &  & \multicolumn{3}{c@{}}{LSUN}\\
	\cmidrule(l{0.7em}){3-5}
  {Method} & {CelebA-HQ}&{Bedroom} & {Tower} & {Church} \\
	\midrule
 SPN \cite{MenickK19}  & \bfseries 0.61 & {--} & {--} & {--} \\
 Glow \cite{KingmaD18} & 1.03  & 1.20 & {--} & {--} \\
 MaCoW \cite{MaKZH19}  & 0.67 & 0.98 & 1.02 & 1.09 \\
 WaveletFlow \cite{YuDB20}  & 0.94 & {--} & {--} & {--}\\
	\midrule
	PixelPyramids \emph{(ours)} &  \bfseries 0.61  & \bfseries 0.88  & \bfseries 0.95 & \bfseries 1.07 \\
	\bottomrule
  \end{tabularx}
  \caption{Evaluation on the 5-bit CelebA-HQ $(256 \times 256)$ and LSUN $(128 \times 128)$  datasets in bits/dim (lower is better).}
  \label{tab:celeba_lsun}
  \vspace{-0.5em}
\end{table}

\section{Experiments}
To show the advantages of our PixelPyramids for density estimation and high-resolution image synthesis, we evaluate on the CelebA-HQ ($256 \times 256$,  $1024 \times 1024$) \cite{KarrasALL18}, LSUN -- bedroom, church, and tower  ($128 \times 128$) \cite{YuZSSX15}, and ImageNet ($128 \times 128$) \cite{Russakovsky:2015:ILS} datasets.
We report the density estimates as per-pixel likelihoods in terms of bits per dimension \cite{OordKK16}. 
Additionally, we provide details on the computational efficiency of our model in comparison to competing methods in terms of training and sampling speed.

\subsection{Implementation details} 
\label{sec:experiments-archi}
The number of pyramid levels ($L$) are chosen such that the coarsest scale has a spatial resolution of $4 \times 4$.
Thus, $L=16$ for CelebA-HQ ($1024 \times 1024$).
The U-Net applied to the fine components consists of two downsampling and two upsampling layers at level $L-1$; an additional down/upsampling layer is added as we move to higher resolutions in the pyramid.
Two squeeze operations were found to be sufficient to model local dependencies at levels $\{1,\ldots,L-1\}$ in all experiments (see \cref{tab:ablation}).
The Conv-LSTM consists of 2 layers with 128 channels and $3 \times 3$ filters.
The autoregressive model at the coarsest level is a PixelCNN++ \cite{SalimansK0K17} with 96 channels and 5 residual layers.
The number of mixture components across datasets is set to 10.
The model is trained with the Adam optimizer \cite{KingmaB14} with an initial learning rate of $10^{-3}$. 
The batch size is set to 64 for all datasets except for CelebA-HQ $1024 \times 1024$, where it is 32.
For the high-resolution setting, \ie $1024 \times 1024$, we use patchwise training \cite{YuDB20}. This is enabled by our fully convolutional 
model (see supplemental material).
Furthermore, the per-pixel values of the bimodal distribution of the fine components (\cf \cref{fig:hist_marginal}) with its cyclic structure are shifted by 128 before encoding with \cref{eq:per-pixel-conditionals}, resulting in a unimodal distribution and providing for a better model fit.

\begin{figure}[t]
        \centering
        \includegraphics[width=0.95\linewidth]{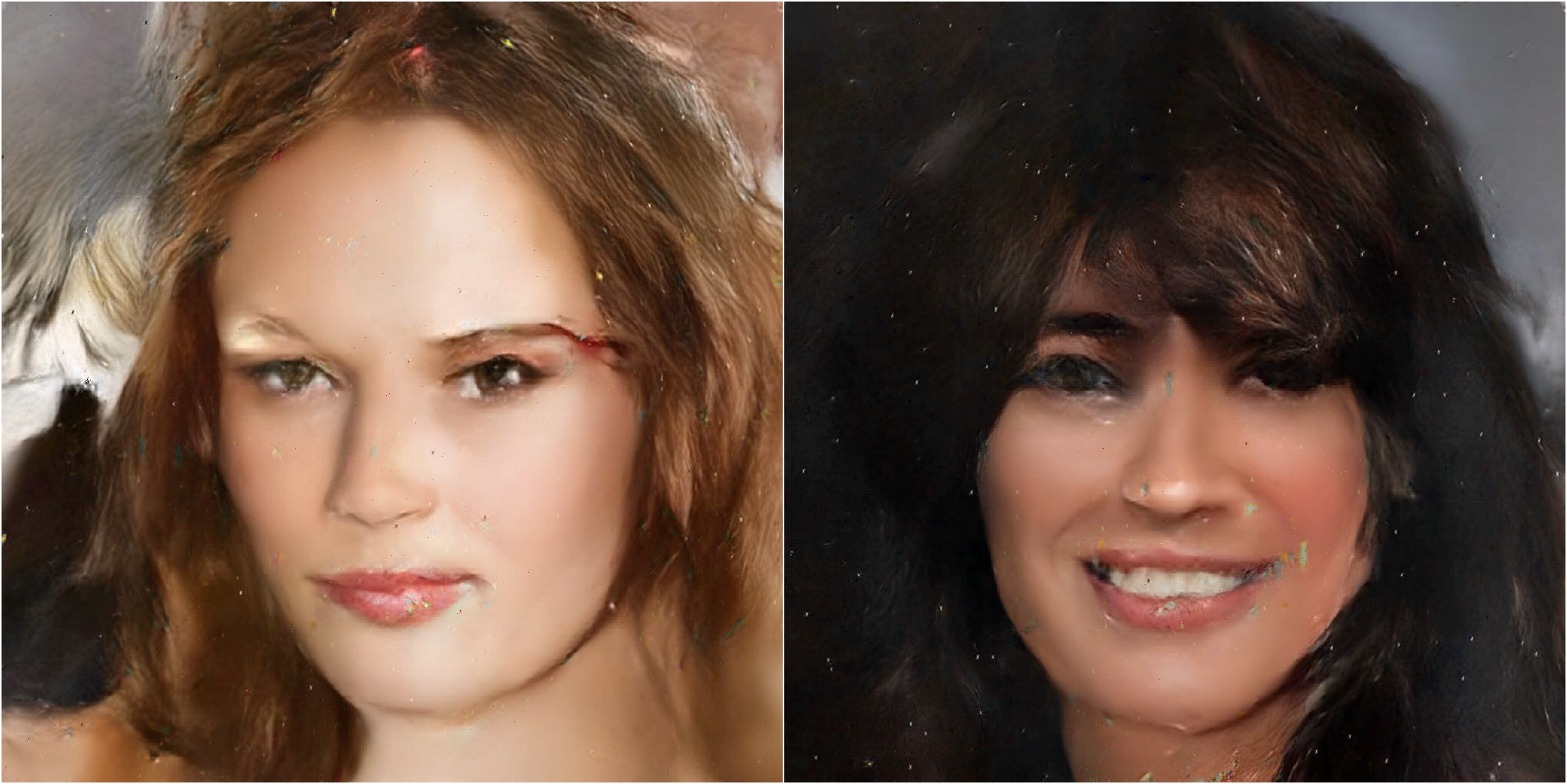}
\caption{Random samples from 8-bit CelebA-HQ ($1024 \times 1024$).}
\label{fig:celeba1024_samp}
  \vspace{-0.5em}
\end{figure}

\begin{table}[t]
  \centering
  \scriptsize
  \begin{tabularx}{\columnwidth}{@{}X@{\hskip 1em}S[table-format=1.2]@{}}
	\toprule
    {Method} & {bits/dim. $(\downarrow)$}\\
	\midrule
	ParallelWavelet \cite{KingmaD18} & 3.55 \\
	SPN \cite{MenickK19}  & \bfseries 3.08 \\
	\midrule
	PixelPyramids \emph{(ours)} & 3.40  \\
	\bottomrule
  \end{tabularx}
  \caption{Evaluation on 8-bit ImageNet ($128 \times 128$) dataset.}
  \label{tab:imagenet}
  \vspace{-0.5em}
\end{table}

\begin{figure*}[t]
\centering
	\begin{subfigure}[b]{0.5\textwidth}
        \centering
        \includegraphics[width=0.9\linewidth]{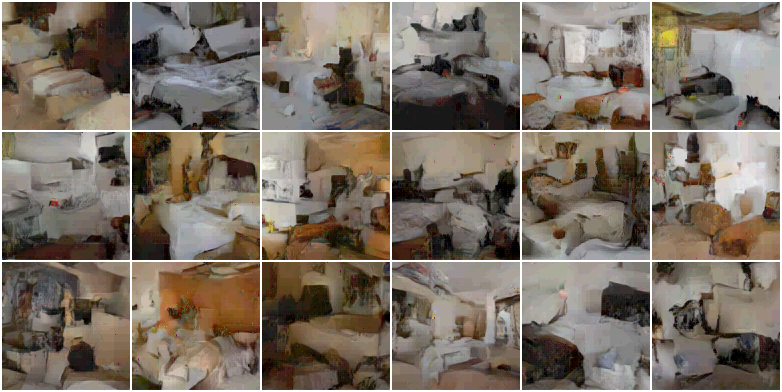}
        \caption{Random samples from 5-bit LSUN Bedroom ($128 \times 128$).}
        \label{fig:lsunbed128_samp}
    \end{subfigure}\hfill%
    \begin{subfigure}[b]{0.5\textwidth}
        \centering
        \includegraphics[width=0.9\linewidth]{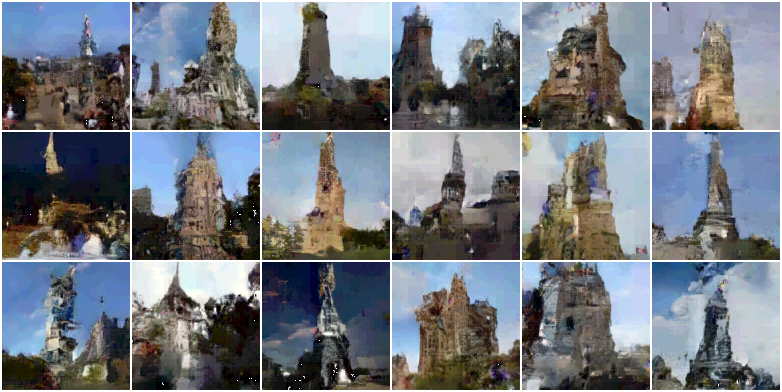}
       \caption{Random samples from 5-bit LSUN tower ($128 \times 128$).}
        \label{fig:lsuntower128_samp}
    \end{subfigure}
    \begin{subfigure}[b]{0.5\textwidth}
        \centering
        \includegraphics[width=0.9\linewidth]{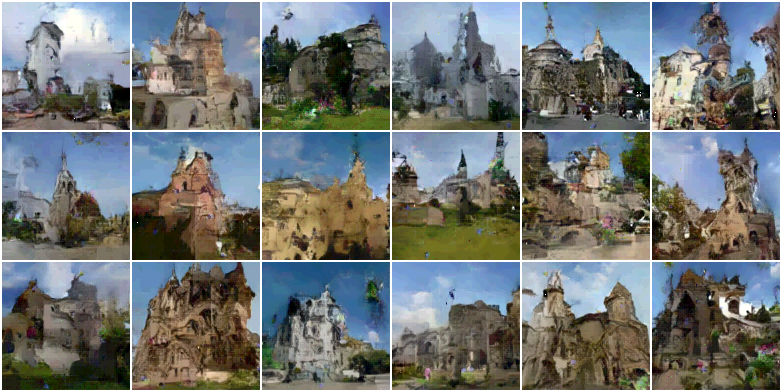}
        \caption{Random samples from 5-bit LSUN church outdoor ($128 \times 128$).}
        \label{fig:lsunchurch128_samp}
    \end{subfigure}%
    \begin{subfigure}[b]{0.5\textwidth}
        \centering
        \includegraphics[width=0.9\linewidth]{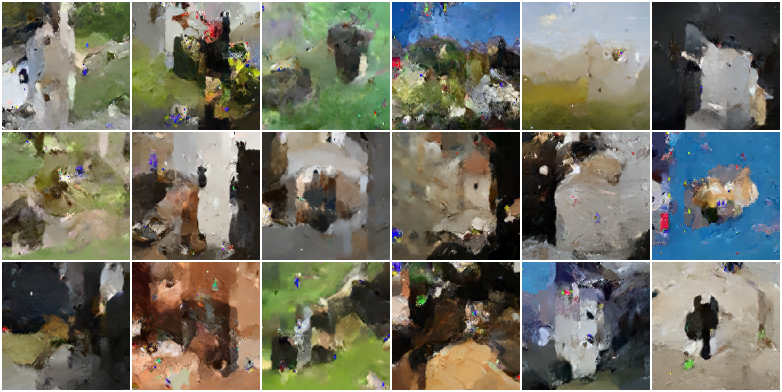}
        \caption{Random samples from 8-bit ImageNet ($128 \times 128$).}
       \label{fig:imagenet128_samp}
    \end{subfigure}
\caption{Random samples from our \emph{PixelPyramids} model on different datasets.}
\label{fig:samp}
\vspace{-0.5em}
\end{figure*}

\subsection{Density estimation}
We compare our approach for high-resolution images against models with state-of-the-art performance on challenging high-resolution datasets, the autoregressive Subscale Pixel Networks (SPN) \cite{MenickK19} and flow-based models Glow \cite{KingmaD18}, Masked Convolutional Generative Flow (MaCoW) \cite{MaKZH19}, and WaveletFlow \cite{YuDB20}.
The number of bits per color channel for each setting is chosen to be consistent with prior work \cite{KingmaD18,MenickK19,YuDB20}.
For CelebA-HQ ($256 \times 256$) and LSUN (bedroom, tower, and church outdoor; $128 \times 128$), the experiments are performed on 5-bit images.

As \cref{tab:celeba_lsun} shows, our PixelPyramids outperform the flow-based models applied directly in image space, Glow and MaCoW, \eg~on CelebA-HQ by $\sim\!40.7\,\%$ and  $\sim\!8.9\,\%$, respectively.
The improvement in density estimation is consistent across datasets and illustrates the high representational power of our framework over standard flow models. 
Moreover, our approach outperforms WaveletFlow with its scale-space representations applied to flow by $\sim\!35.1\,\%$. 
This highlights the advantage of our easier-to-model low-entropy Paired Pyramid representation.
Our PixelPyramids close the performance gap to fully autoregressive models like SPN despite their partial autoregressive structure.

\begin{table}[t]
  \centering
  \scriptsize
  \begin{tabularx}{\columnwidth}{@{}X@{\hskip 1em} S[table-format=1.2]@{}}
	\toprule
	{Method} & {bits/dim. $(\downarrow)$} \\
	\midrule
    WaveletFlow \cite{YuDB20} & 1.34 \\
    \midrule
	PixelPyramids \emph{(ours)} &  \bfseries 0.58 \\
	\bottomrule
  \end{tabularx}
  \caption{Evaluation on 8-bit CelebA-HQ ($1024 \times 1024$).}
  \label{tab:celeba256}
  \vspace{-0.5em}
\end{table}

\cref{tab:imagenet} further compares our approach on ImageNet $128 \times 128$ (8 bits).
Our block-autoregressive PixelPyramids yield better density estimates compared to the multi-scale ParallelWavelet, which applies a PixelCNN at every pyramid level to the sub-images in pixel space. 
This shows the advantage of our Paired Pyramid representation, including that it can be modeled well with only a partially autoregressive structure of the Conv-LSTM (see also \cref{tab:ablation}). 
While SPNs still show better density estimates for this dataset, they are fully autoregressive in the spatial dimensions as well as for the image values, making them not scalable.

One of the main advantages of our approach is its applicability to high-resolution images. 
WaveletFlow is the only previous work based on exact inference that reports density estimates on 8-bit images of CelebA-HQ ($1024 \times 1024$).
PixelPyramids yield \emph{a new state of the art} where density estimates are improved to $\sim\!\! 44\,\%$ of the bits/dim obtained with WaveletFlow, \ie from 1.34 bits/dim to 0.58 bits/dim (\cref{tab:celeba256}).
This large gain can be attributed to the ease of modeling the fine components from the Paired Pyramid representation compared to the scale-specific representations from the Haar wavelet.
Moreover, at every level the Conv-LSTMs model the local structure in the scale-specific representations of our approach compared to WaveletFlow, which builds upon non-autoregressive models like Glow.

Note that the Paired Pyramid and consequently PixelPyramids have limited applicability to low-resolution images, \eg CIFAR10 \cite{krizhevsky2009learning}, as they yield fine components with similar (high) entropy as the original image (see supp.~mat.).

\begin{table}[t]
  \centering
  \scriptsize
  \begin{tabularx}{\columnwidth}{@{}X@{\hskip 1em}S[table-format=1]@{}}
	\toprule
    {Method} & {bits/dim. $(\downarrow)$}\\
	\midrule
	PixelPyramids & 0.61 \\
	\midrule
	PixelPyramids (no modulo difference)  & 0.76 \\
	PixelPyramids (non-autoregressive, $n_\text{squeeze}=0$) & 0.83\\
	PixelPyramids ($n_\text{squeeze}=1$) & 0.62 \\
	PixelPyramids ($n_\text{squeeze}=2$) & 0.61 \\
	PixelPyramids ($n_\text{squeeze}=3$) & 0.61 \\
	\bottomrule
  \end{tabularx}
  \caption{Model ablations on CelebA ($256 \times 256$).}
  \label{tab:ablation}
  \vspace{-0.65em}
\end{table}

\myparagraph{Ablation studies.}
To validate the advantage of encoding densities with multi-scale low-entropy modulo differences, we compare against a baseline variant without the modulo difference, where the joint pixel distribution is decomposed similar to a regular image pyramid as $p(\mathbf{I}_0)=p(\mathbf{I}_L)\prod_{i=1}^L p(\mathbf{I}'_i \mid \mathbf{I}_i)$; all other model details remain unchanged.
We term this \emph{PixelPyramids (no modulo difference)}.
From \cref{tab:ablation}, we observe a clear drop in performance from 0.61 bits/dim to 0.76 bits/dim, indicating that the images $\mathbf{I}'_i$ have complex structural dependencies, which are difficult to model with a computationally inexpensive partially autoregressive structure.
This shows the benefit of the low-entropy modulo difference for generative modeling.

Furthermore, \emph{PixelPyramids (non-autoregressive)} shows the importance of modeling local spatial dependencies at each level.
To assess this, we use $n_\text{squeeze}=0$ squeeze operations, which effectively disables the Convolutional LSTM.
Without the partial autoregressive structure, the model cannot encode the local dependencies (\cf \cref{fig:mi_stats}), leading to a clear performance drop from 0.61 bits/dim to 0.83 bits/dim.
This highlights that the combination of a low-entropy representation and a partial autoregressive structure is an important design choice for effective density estimation.

\begin{figure}[t]
        \centering
        \includegraphics[width=0.7\columnwidth]{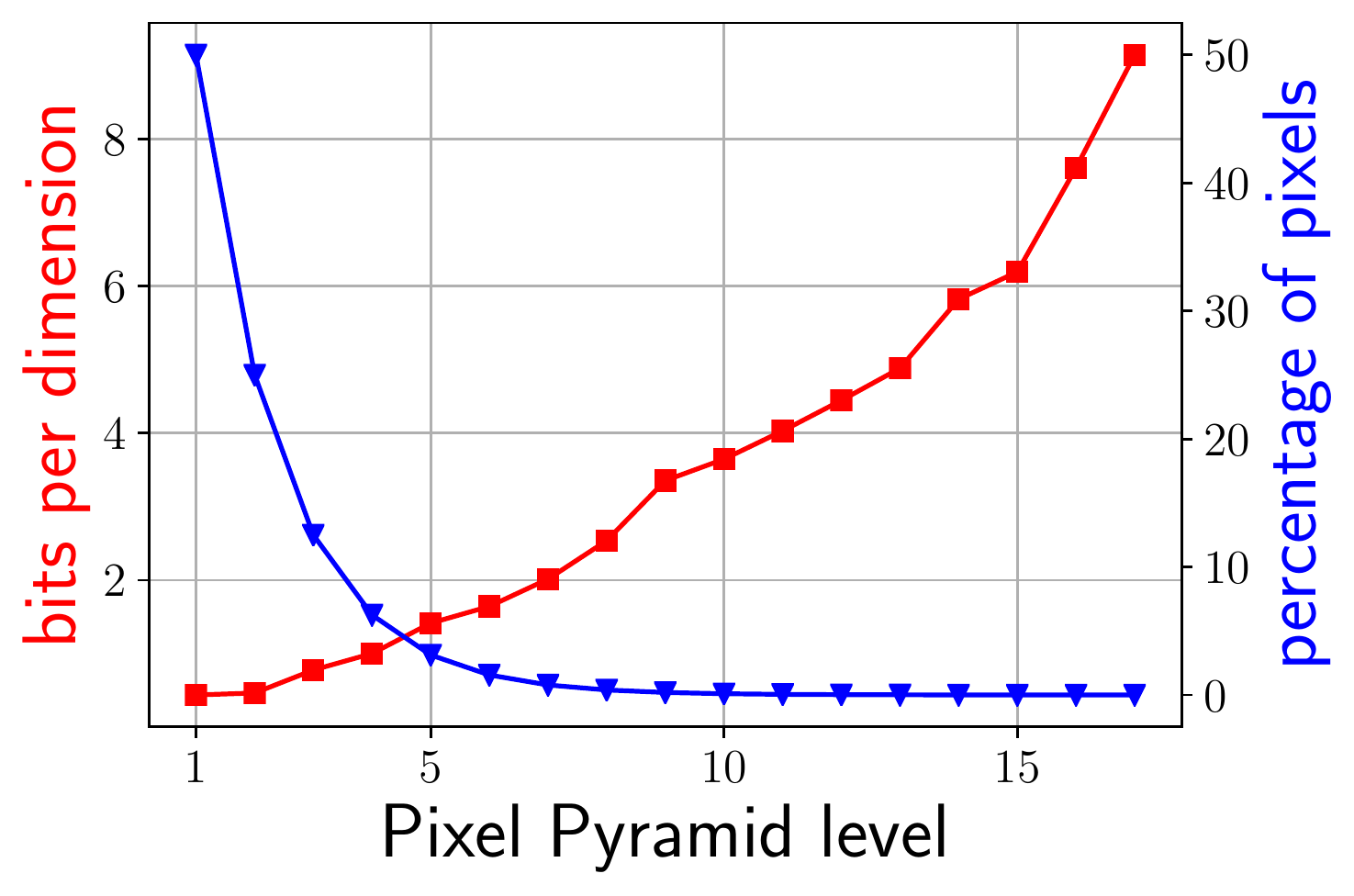}
\caption{\emph{Contribution of each pyramid level} to density estimates under the model \emph{(red)} and the percentage of pixels \emph{(blue)}.}
\label{fig:perlevelbpd}
\vspace{-0.8em}
\end{figure}

Finally, we analyze the effect of varying the number of squeeze operations, $n_\text{squeeze}$, which essentially regulates the sequential dependencies modeled by the Conv-LSTM in our framework.
We observe that the density estimates improve significantly from 0.83 bits/dim to 0.62 bits/dim with a single squeeze operation, and improve further by 0.01 bits/dim with the default of two squeeze operations.
$n_\text{squeeze}=3$ leads to no additional benefit, since $n_\text{squeeze}=2$ already implies a fully-autoregressive modeling at higher pyramid levels with limited spatial resolution, which have the most complex dependency structure (\cf \cref{fig:hist_marginal,fig:perlevelbpd}).
Furthermore, the higher resolution levels containing the vast majority of pixels have low density estimates under the model with $n_\text{squeeze}=2$ compared to pixels at lower resolution levels (\cref{fig:perlevelbpd}).
This demonstrates the sufficiency of modeling local dependencies at higher resolutions in PixelPyramids.

\myparagraph{Image synthesis.}
Qualitative image synthesis results on the 5-bit CelebA ($256 \times 256$, \cref{fig:celeba256_samp}), 8-bit CelebA ($1024 \times 1024$,  \cref{fig:celeba1024_samp}),  5-bit LSUN (bedroom, tower, and church outdoor;  $128 \times 128$), as well as 8-bit ImageNet ($128 \times 128$) datasets (\cref{fig:samp}) show the ability of our PixelPyramids to generate high-quality, diverse samples of high resolution.
Since the sampling step involves the inverse transformation of \cref{eq:inv_mod_diff}, cyclic shifts of pixel values can occur if sampled values fall in the vicinity of 0 or 255.
This leads to occasional pixel-level outliers (also noticeable in \cref{fig:celeba1024_samp} as red salt \& pepper noise).
This could be resolved by replacing outlier pixels with the median of their neighborhood.
We provide additional analyses in the supplemental material.

\myparagraph{Computational cost.} 
The computational efficiency of our model is shown in \cref{tab:comp_cost}.
All the experiments are performed on a single Nvidia TITAN X GPU.
The number of parameters of PixelPyramids is similar to that of Glow and MaCoW. 
Despite having more parameters than WaveletFlow, the training speed of our model is of the same order, owing to its fully convolutional architecture at every level. 
Our block-autoregressive structure provides efficient sampling speeds compared \emph{even} to flow-based models, which already allow for fast sampling.
Moreover, our sampling speeds are \emph{orders of magnitude} faster than fully autoregressive models.
Thus, PixelPyramids provide both a computationally efficient setup \emph{and} accurate density estimation.

\begin{figure}[t]
    \centering
    \includegraphics[width=0.8\columnwidth]{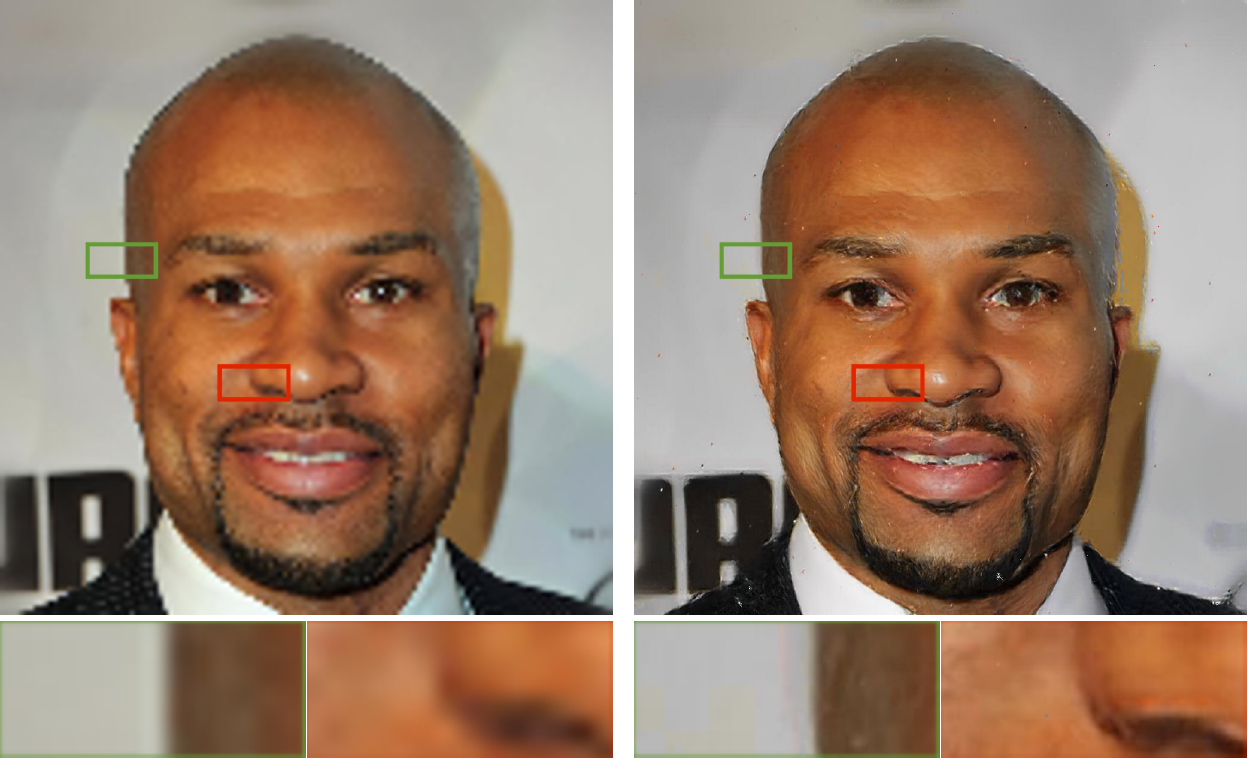}
    \caption{Super-resolution with PixelPyramids resizing a $128 \times 128$ image to $1024 \times 1024$ on the 8-bit CelebA-HQ ($1024 \times 1024$).}
    \label{fig:super_res}
    \vspace{-0.5 em}
\end{figure}

\begin{table}[t]
  \centering
  \scriptsize
 \begin{tabularx}{\columnwidth}{@{}X@{}S[table-format=4]@{\,}s[table-unit-alignment=left]S[table-format=1.3(2)e2,separate-uncertainty=true]S[table-format=1.3(3)e2,separate-uncertainty = true]@{}}
	\toprule
	{Method} & \multicolumn{2}{c}{Params}&{Training Speed} & {Sampling Speed} \\
	\midrule
	PixelCNN++ \cite{SalimansK0K17} & 70 & \si{\mega} & 1.35(10)e2 & 6.00(2)e6  \\
	SPN\footnotemark[3] \cite{MenickK19}  & 50 & \si{\mega} & {--} & {--}  \\
    Glow \cite{KingmaD18}  &171 & \si{\mega} & 1.79(10)e3 & 5.31(34)e2 \\
	MaCoW \cite{MaKZH19}  & 177 & \si{\mega} &  {--} & 4.34(20)e2  \\
	WaveletFlow \cite{YuDB20} & 52 & \si{\mega} & 1.47(20)e1 & 9.98(86)e2\\
	\midrule
	PixelPyramids & 166 & \si{\mega} & 1.98(16)e1 & 7.00(10)e1\\
	\bottomrule
  \end{tabularx}
  \caption{Comparison of training and sampling speeds (ms/image) on CelebA-HQ ($256 \times 256$) with 100 random seeds.}
  \label{tab:comp_cost}
  \vspace{-0.75em}
\end{table}
\footnotetext[3]{The sampling speed of SPN could not be reported due to the nonavailability of an official implementation. Therefore, the sampling speed of PixelCNN++ with 0.67 bits/dim and same order of sampling speed (our implementation reproducing results on CIFAR10) is included as reference.}

\myparagraph{Super-resolution.}
Similar to WaveletFlow \cite{YuDB20}, we show qualitatively the applicability of PixelPyramids for super-resolution without any explicit task-specific learning (\cref{fig:super_res}). 
Given the low-resolution coarse image, the fine components are iteratively generated to obtain the high-resolution output.
The integration of PixelPyramids with methods for probabilistic super-resolution \cite{LugmayrDGT20} is an interesting direction of future research.
\section{Conclusion}
We present PixelPyramids, a novel and computationally efficient block-autoregressive framework for estimating densities and synthesis of high-resolution images.
We show that multi-scale lossless representations obtained from Paired Pyramids, consisting of low-entropy fine components, provide for an easier to model localized dependency structure. 
This makes exact inference computationally efficient through partial autoregressive connections.
Our framework provides a promising direction for extending exact inference models to high-resolution datasets. 

{\small \myparagraph{Acknowledgement.} SM and SR acknowledge the support by the German Research Foundation as part of the Research Training Group Adaptive Preparation of Information from Heterogeneous Sources (AIPHES) under grant No.~GRK 1994/1.
Further, this work has in part received funding by the European Research Council (ERC) under the European Union’s Horizon 2020 research and innovation programme (grant agreement No.~866008).}

{\small
\bibliographystyle{ieee_fullname}
\bibliography{egbib}
}

\clearpage
\appendix
\appendixpage
\addappheadtotoc

We provide the proof of Lemma 3.1 in the main paper, as well as additional details on the datasets and the implementation. 
We further include additional qualitative examples.
\section{Proof of Lemma 3.1}
\paragraph{Lemma 3.1.}
\emph{Let the sampled image $\mathbf{I}$ be of resolution $\left[N_0,N_0,C\right]$, then the worst-case number of steps $T$ (length of the critical path) required is $\mathcal{O}(\log N_0)$.}
\begin{proof}
At the first sampling step, \ie at the coarsest level, the spatial dimensionality of the image is $\left[N_0/ 2^{\lfloor \left(L+1\right)/2 \rfloor}, N_0/2^{ \lfloor L/2 \rfloor}, C\right]$.
At each level starting from the coarsest, the spatial resolution increases by a factor of 2, alternatingly along the rows and columns. 
Thus, to generate an image of size $[N_0, N_0,  C]$ starting from an image of size $\left[1, 1, C\right]$, the number of levels of the coarse-to-fine pyramid to traverse equals 
\begin{align}
    L = 2\cdot\log N_0 .
\end{align}
At the coarsest level of spatial resolution $1 \times 1$, with its fully autoregressive PixelCNN structure, the number of sampling steps is equal to the spatial dimension and, therefore, is given as
\begin{align}
    T_L = 1.
\end{align}
Let the number of squeeze operations applied at level $i$ be $n_{S_i}$ such that we obtain $4^{n_{S_i}}$ subsampled images from $\mathbf{F}_i$. 
This implies that the number of sequential steps at level $i$ is
\begin{align}
    T_i = 4^{n_{S_i}},
\end{align}
assuming that each autoregressive sampling step can be carried out in parallel (hence in constant time).
Therefore, the total number of sequential steps (length of the critical path) required for sampling is
\begin{align}
    \begin{split}
        T &= \sum_{i=1}^{L} T_{i} = T_{L}  + \sum_{i=1}^{L-1} T_{i} \\
         & = 1 + \sum_{i=1}^{L-1} 4^{n_{S_i}}\\
    \end{split}
\end{align}
Under the assumption that the number of squeeze operations at any level is constant with $\mathcal{O}(1)\ni 4^{n_{S_i}}\ll N_0$, we obtain the the number of sampling steps required as 
\begin{align}
        T \in 2\cdot\log N_0 \cdot \mathcal{O}(1) = \mathcal{O}(\log N_0)
\end{align}
\end{proof}

\section{Additonal Implementation Details}

\myparagraph{Datasets.} CelebA-HQ \cite{KarrasALL18} consists of 30K images of which 26K are used for training, 1000 images for validation, and 3000 images are provided in the test set.

The LSUN \cite{YuZSSX15} bedroom, church outdoor, and tower datasets consist of 3M, 126K, and 700K images, respectively, in the training set.
The validation set consists of 300 images for each of the datasets.
For training at $128 \times 128$ resolution, similar to \cite{KingmaD18,MaKZH19}, the images are first center cropped to the spatial resolution of $256 \times 256$ and then resized to a spatial resolution of $128 \times 128$.

Further, ImageNet \cite{Russakovsky:2015:ILS} consists of 1.3M images in the training set and 50K images in the test set.
We follow \cite{MenickK19,ReedOKCWCBF17} for resizing the images to size $128 \times 128$,  where the images are first cropped along the longer spatial dimension and then resized to the desired spatial resolution.

\myparagraph{Optimization.} Similar to \cite{MaKZH19,SalimansK0K17}, we use Adam \cite{KingmaB14}
with an initial learning rate of $10^{-3}$. Additionally, the parameters are taken as $\beta_1=0.95$, $\beta_2 =0.9995$ and Polyak averaging is set to 0.9995. 
The learning rate decays exponentially at a rate of 0.999995.

\begin{figure}[t]
    \centering
   \includegraphics[width=0.75\columnwidth]{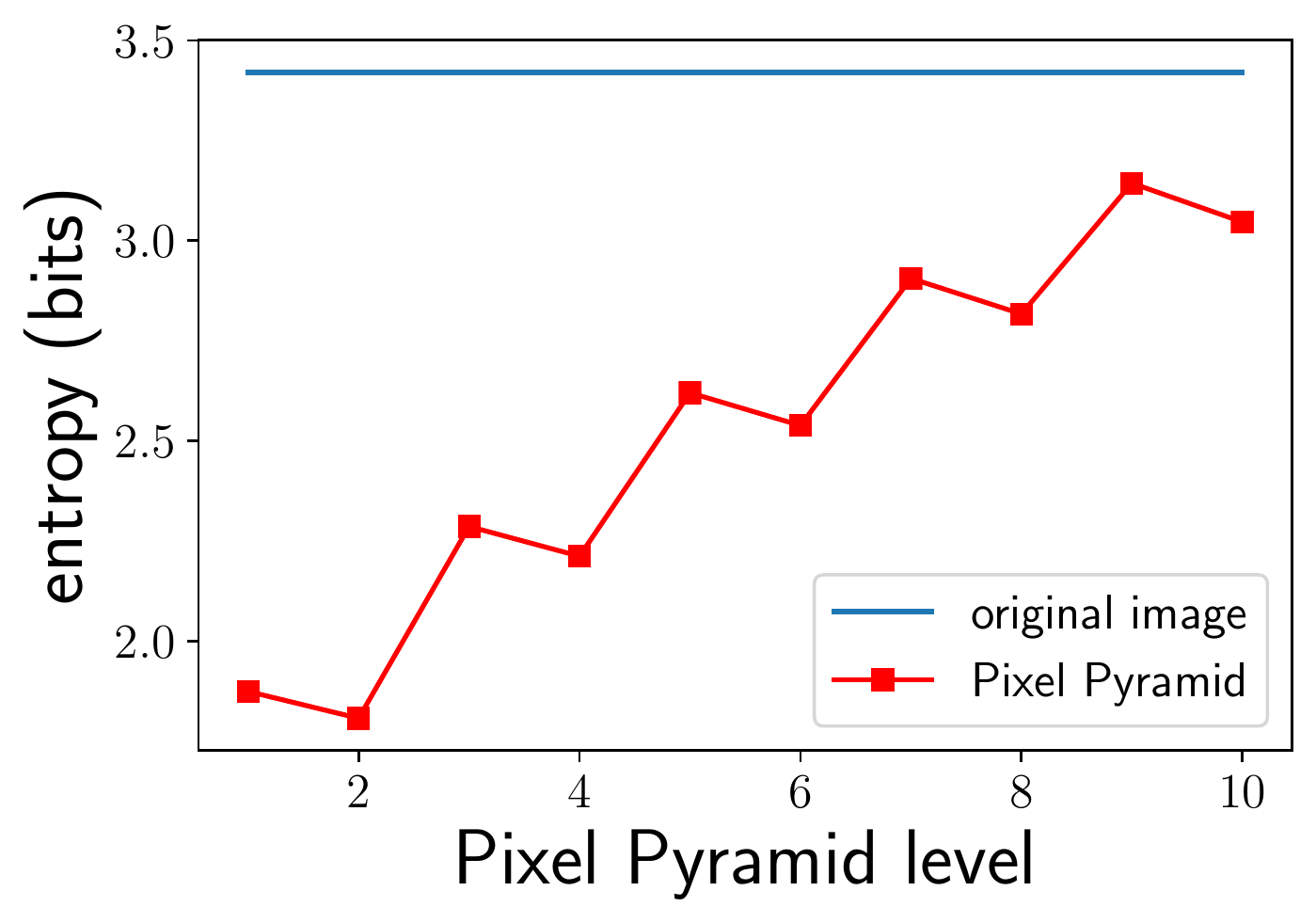}
    \caption{\emph{Low-entropy decomposition.} Entropy values at different pyramid levels on ImageNet ($128 \times 128$). In contrast to the other datasets considered in the main paper, the entropy values for the decomposition of ImageNet are not significantly reduced compared to the original image.}
    \label{fig:imagenet_entropy}
    \vspace{-0.5em}
\end{figure}

\begin{table}[h]
  \centering
  \footnotesize
 \begin{tabularx}{\columnwidth}{@{}Xl@{}}
	\toprule
	Dataset & Parameter Count\\
	 \midrule
	CelebA-HQ ($256 \times 256$) & 166M \\
	CelebA-HQ ($1024 \times 1024$) & 349M \\
	LSUN ($128 \times 128$) & 224M\\
	ImageNet ($128 \times 128$) & 346M\\
	\bottomrule
  \end{tabularx}
  \caption{Parameter count of PixelPyramids across different datasets.}
  \label{tab:params_suppmat}
  \vspace{-0.5em}
\end{table}
In \cref{tab:params_suppmat} we provide the number of parameters for different datasets. The number of parameters varies depending on the size and the complexity of the dataset. 
Within the U-Net module at each level, the number of channels increases by a factor of two for every downsamling/upsampling layer. The first layer has 64 channels for the CelebA-HQ ($256 \times 256$, $1024 \times 1024$) and LSUN ($128 \times 128$) datasets.
Owing to the multimodality of the dataset, ImageNet ($128 \times 128$) even at $128 \times 128$ resolution requires same number of parameters as CelebA-HQ $1024 \times 1024$, as we need to increase the width of the first layer of the U-Net module to 128 channels.
This design choice is further supported by the Paired Pyramid decomposition of ImageNet $128 \times 128$ (\cref{fig:imagenet_entropy}), which shows that the entropy is not reduced as significantly for the fine components as is the case for the CelebA-HQ dataset (\cf~\cref{fig:hist_marginal}).\footnotemark[4]\footnotetext[4]{The entropy values across different datasets are not comparable due to different preprocessing procedures.}

\section{Additional Results}
\begin{figure}[t]
    \centering
   \includegraphics[width=0.75\columnwidth]{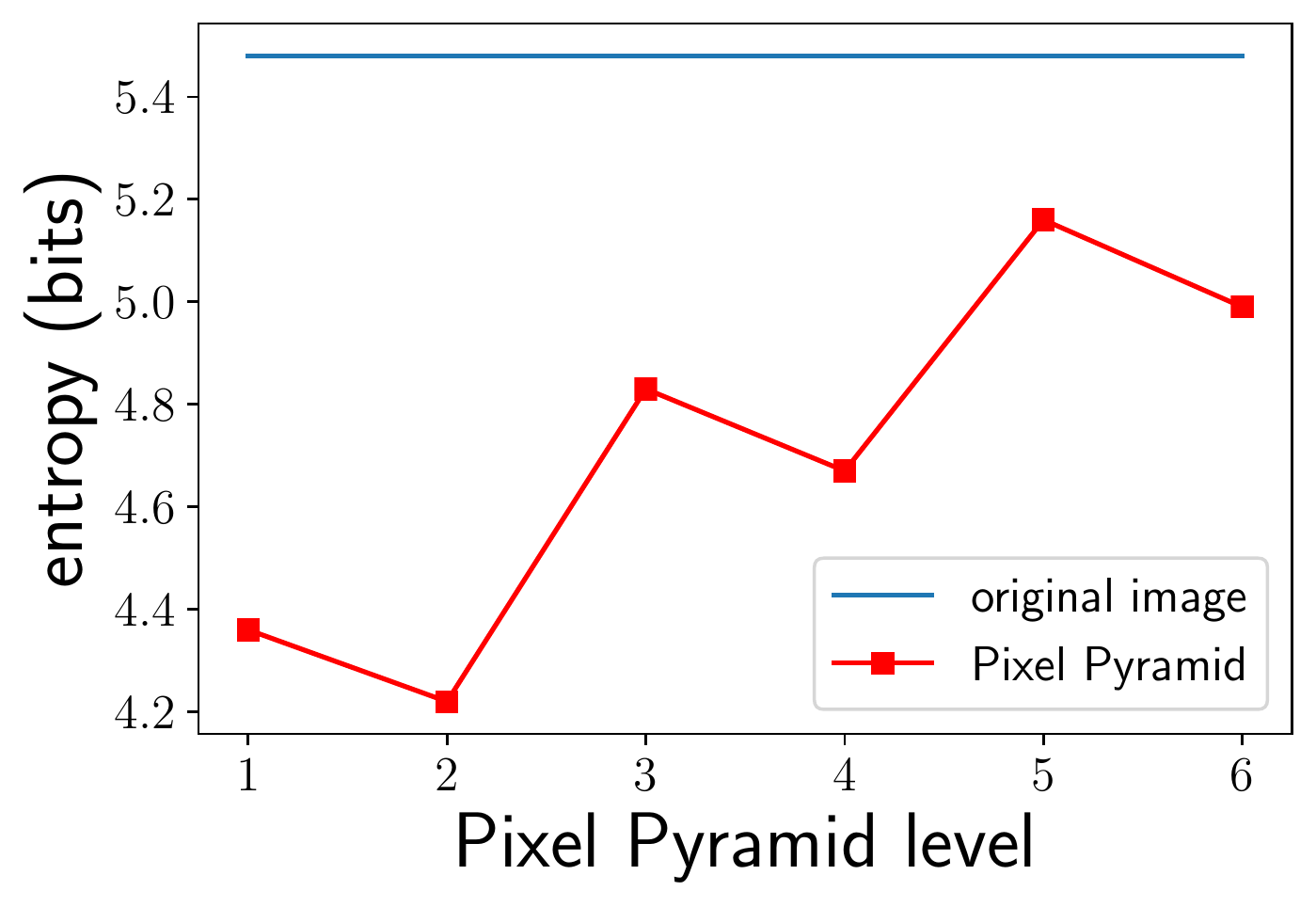}
    \caption{\emph{Low-entropy decomposition.} Entropy values at different pyramid levels on CIFAR10. The entropy values for the decomposition are not significantly reduced compared to the original image for this multimodal low-resolution dataset.}
    \label{fig:cifar_entropy}
    \vspace{-0.5em}
\end{figure}

In \cref{fig:cifar_entropy} we show the entropy values for the fine components of the Paired Pyramid representation of the low-resolution CIFAR10 \cite{krizhevsky2009learning} dataset. 
In comparison to the CelebA-HQ dataset (\cf~\cref{fig:hist_marginal}), where the entropy values are significantly reduced, \eg for $\mathbf{F}_1$ from  5.51 bits for the original pixel values to 3.14 bits for the fine component, the entropy of the fine components for the CIFAR10 dataset does not reduce significantly, even less so than for ImageNet ($128 \times 128$).
This limits the applicability of our PixelPyramids framework to such multimodal low-resolution datasets, since the fine components of the Paired Pyramid representation have similarly high entropy as the original image.
This is further observed in \cref{tab:cifar10}, where the density estimates on CIFAR10 with fully autoregressive approaches are better compared to that with PixelPyramids with its partial autoregressive structure.    
We thus focus on high-resolution images here, since this is where the key limitations of existing exact inference models currently lie.

\begin{table}[t]
  \centering
  \footnotesize
  \begin{tabularx}{\columnwidth}{@{}X@{\hskip 1em}S[table-format=1.2]@{}}
	\toprule
    {Method} & {bits/dim. $(\downarrow)$}\\
	\midrule
	PixelCNN \cite{OordKEKVG16} & 3.03 \\
	PixelCNN++ \cite{SalimansK0K17}  & 2.92 \\
	Glow \cite{KingmaD18}  & 3.35 \\
	Flow++ \cite{HoCSDA19} & 3.08\\
	Residual Flow \cite{BehrmannGCDJ19} & 3.28\\
	MaCoW \cite{MaKZH19}  & 3.16 \\
	\midrule
	PixelPyramids \emph{(ours)} & 3.19 \\
	\bottomrule
  \end{tabularx}
  \caption{Evaluation on the 8-bit CIFAR10 ($32 \times 32$) dataset.}
  \label{tab:cifar10}
  \vspace{-0.5em}
\end{table}

\section{Additional Examples}
\begin{figure*}[h]
    \centering
    \begin{subfigure}[t]{\textwidth}
     \centering
   \includegraphics[width=0.8\linewidth]{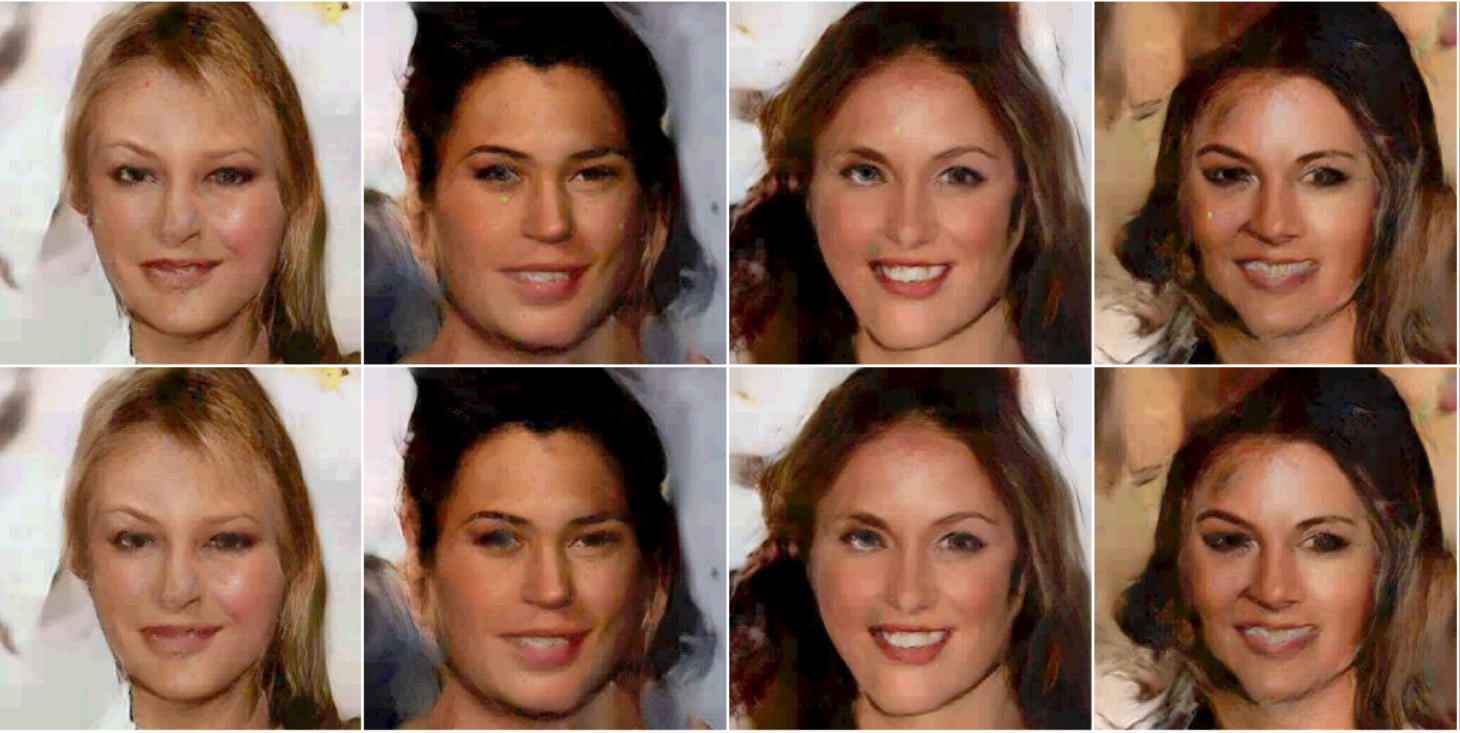}
   \end{subfigure}
   \begin{subfigure}[t]{\textwidth}
    \centering
   \includegraphics[width=0.8\linewidth]{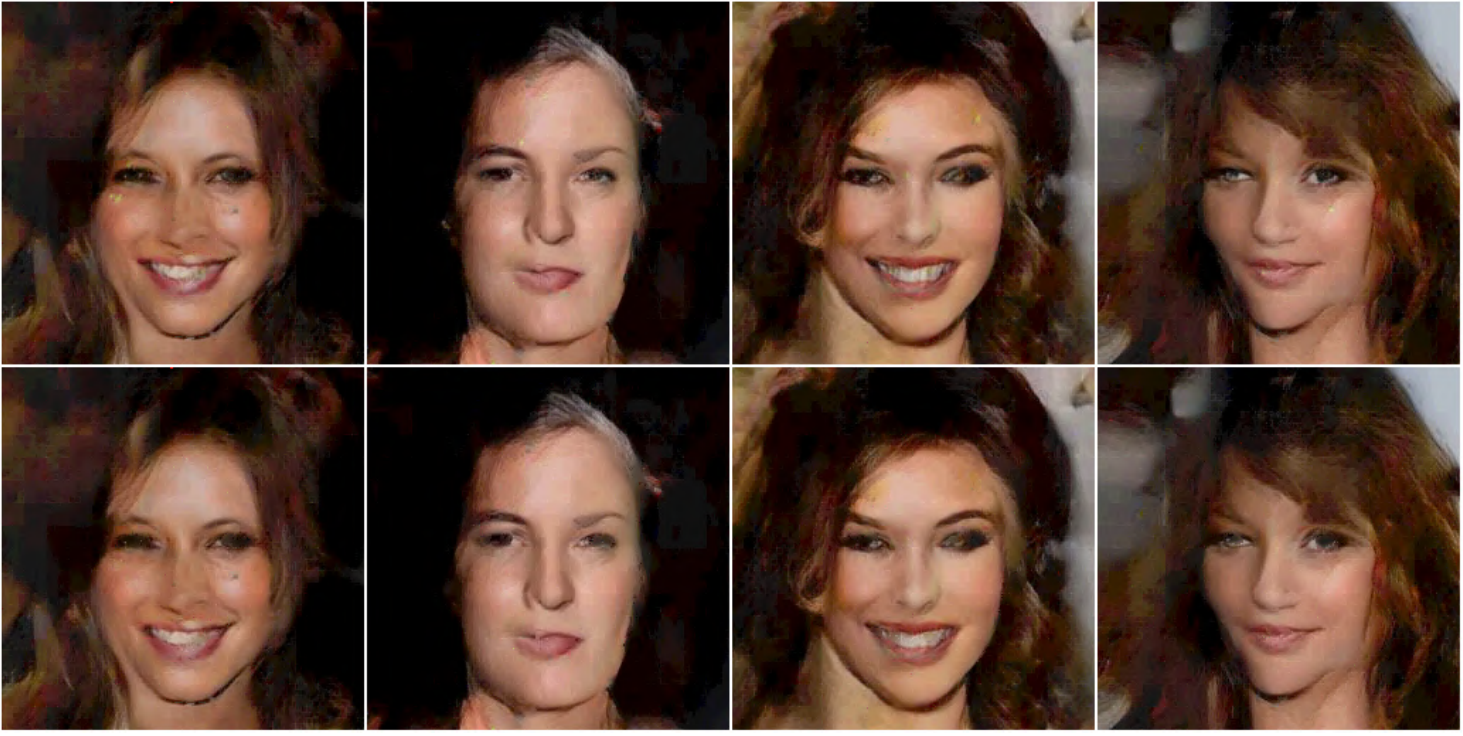}
   \end{subfigure}
    \caption{\emph{Removal of pixel outliers from images synthesized with PixelPyramids on  5-bit CelebA-HQ ($256 \times 256$):} Images generated with our PixelPyramids framework (\emph{row 1~\&~3}); generated images after the application of pixel outlier removal using a median filter (\emph{rows 2~\&~4}, see text).}
    \label{fig:postprocess_suppmat}
    \vspace{-0.5em}
\end{figure*}

To show that the pixel outliers in the  images synthesized with PixelPyramids can indeed be resolved, we include \cref{fig:postprocess_suppmat}, where we apply a median filter to remove the artifacts resulting from the cyclic shift of pixel values in the vicinity of 0 and 255 (\cref{algo:post_samp}).
The pixel outliers are detected using the Isolation Forest algorithm \cite{LiuTZ08} in the HSV space and the outlier pixels are replaced using a median filter over an $7 \times 7$ neighborhood.

\begin{algorithm}[h]
 Sample $\hat{\mathbf{I}} \sim p_{\theta, \phi_L}(\mathbf{I}_{0})$ \tcp*{Sample the image}\
 $\hat{\mathbf{I}} _{\text{med}} \gets {\medianfilter}(\hat{\mathbf{I}}, m)$\tcp*{Median filter}\
 $\hat{\mathbf{I}} _{\text{hsv}} \gets {\rgbtohsv}(\hat{\mathbf{I}} )$\tcp*{Convert to HSV}\
 $\mathbf{O}_{\text{mask}} \gets {\isolationforest}(\hat{\mathbf{I}} _{\text{hsv}}, n)$\tcp*{Outliers}
 $\hat{\mathbf{I}}[\mathbf{O}_{\text{mask}}] \gets \hat{\mathbf{I}}_{\text{med}}[\mathbf{O}_{\text{mask}}]$\tcp*{Assign median}
 
 \caption{Improvement of images synthesized with PixelPyramids using pixel outlier detection. $m$ is the filter size for the median filter and $n$ is the pixel neighborhood for outliers.}
 \label{algo:post_samp}
\end{algorithm}

\begin{figure*}[t]
    \centering
   \includegraphics[width=\textwidth]{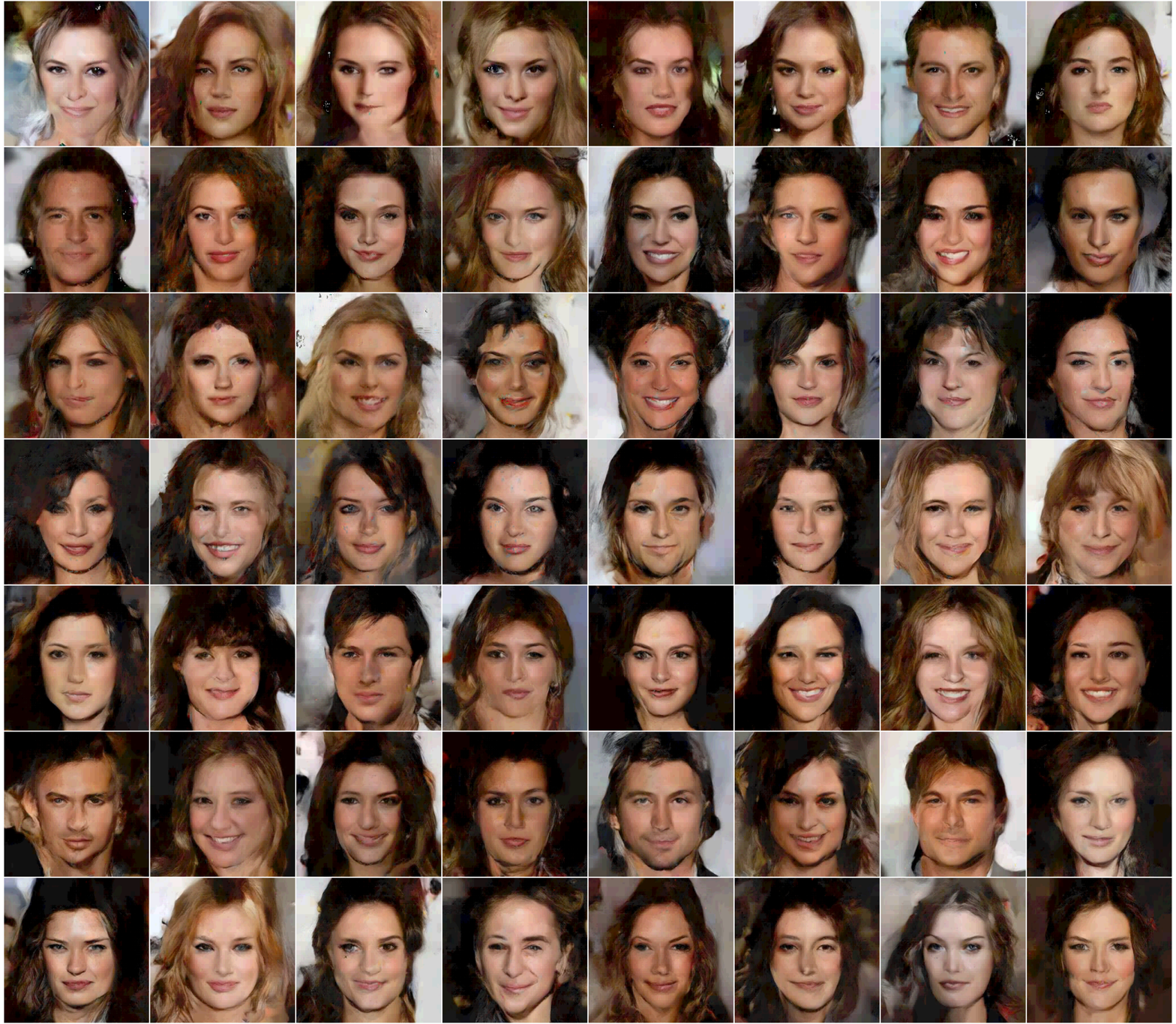}
    \caption{Random samples from 5-bit CelebA-HQ ($256 \times 256$).}
    \label{fig:celeba_suppmat}
    \vspace{-0.5em}
\end{figure*}

\begin{figure*}[t]
    \centering
   \includegraphics[width=\textwidth]{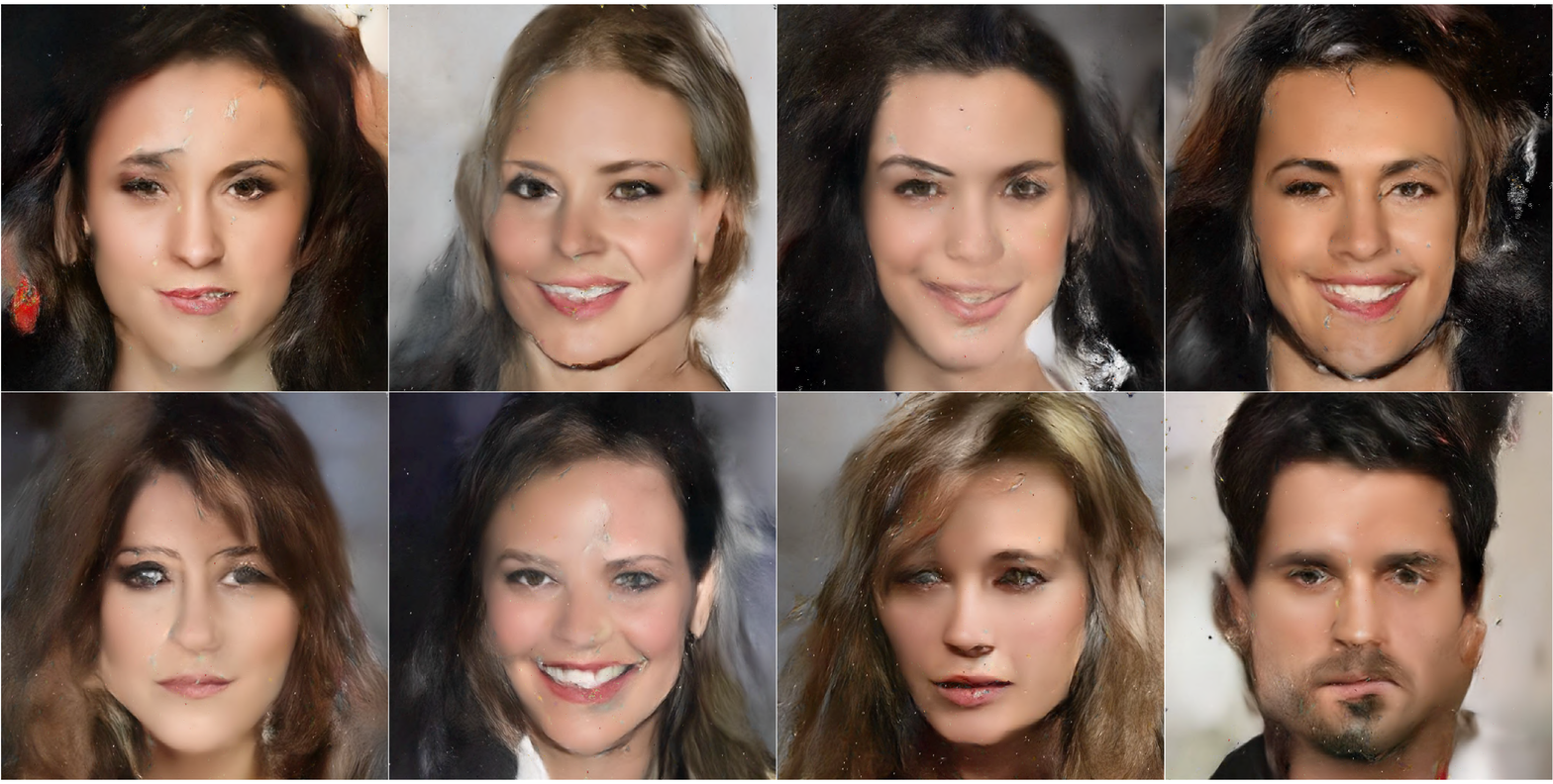}
    \caption{Random samples from 8-bit CelebA-HQ ($1024 \times 1024$).}
    \label{fig:celeba1024_suppmat}
    \vspace{-0.5em}
\end{figure*}

\begin{figure*}[t]
    \centering
   \includegraphics[width=\textwidth]{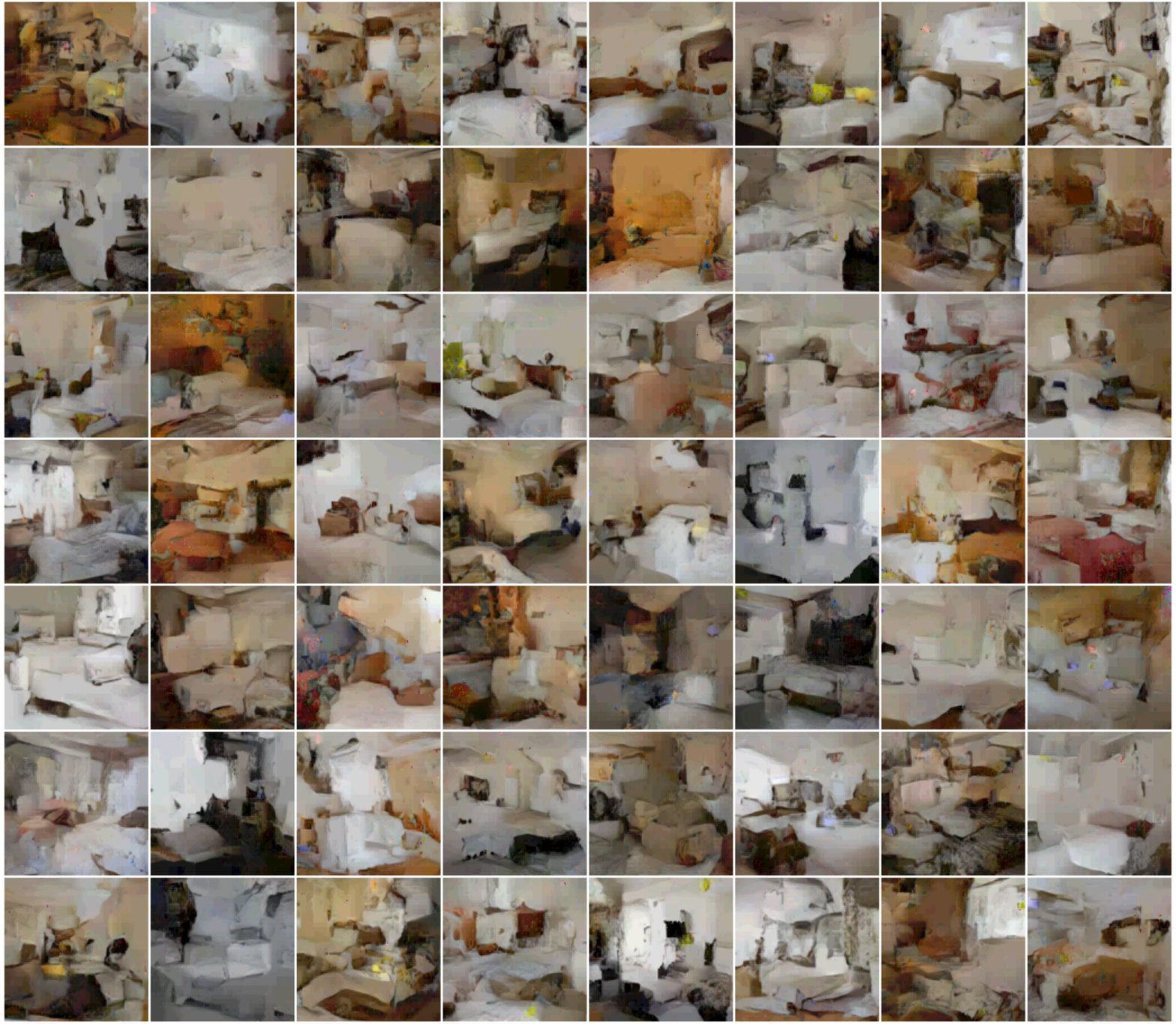}
    \caption{Random samples from 5-bit LSUN bedroom ($128 \times 128$).}
    \label{fig:lsunbedroom_suppmat}
    \vspace{-0.5em}
\end{figure*}

\begin{figure*}[t]
    \centering
   \includegraphics[width=\textwidth]{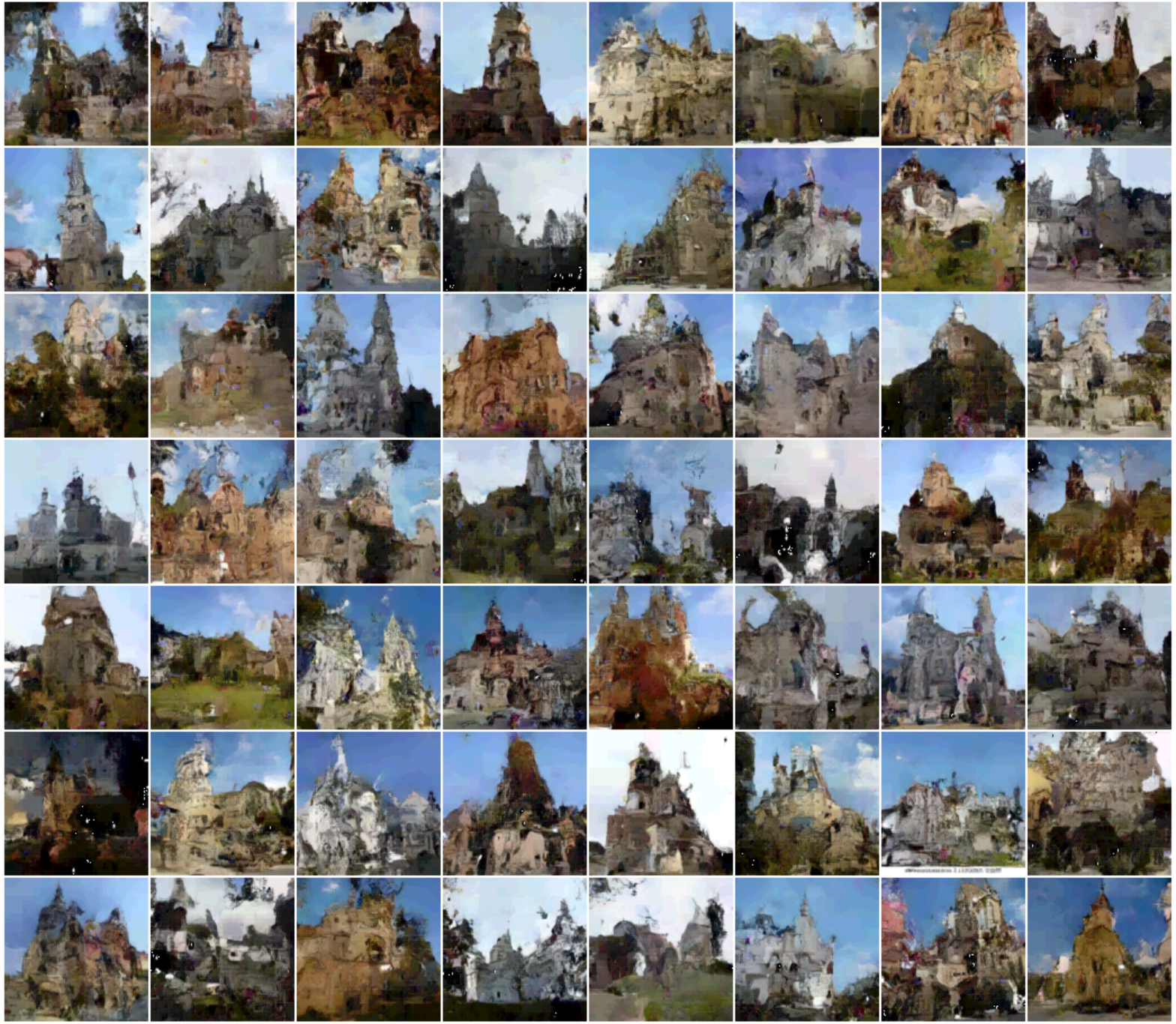}
    \caption{Random samples from 5-bit LSUN church ($128 \times 128$).}
    \label{fig:lsunchurch_suppmat}
    \vspace{-0.5em}
\end{figure*}

\begin{figure*}[t]
    \centering
   \includegraphics[width=\textwidth]{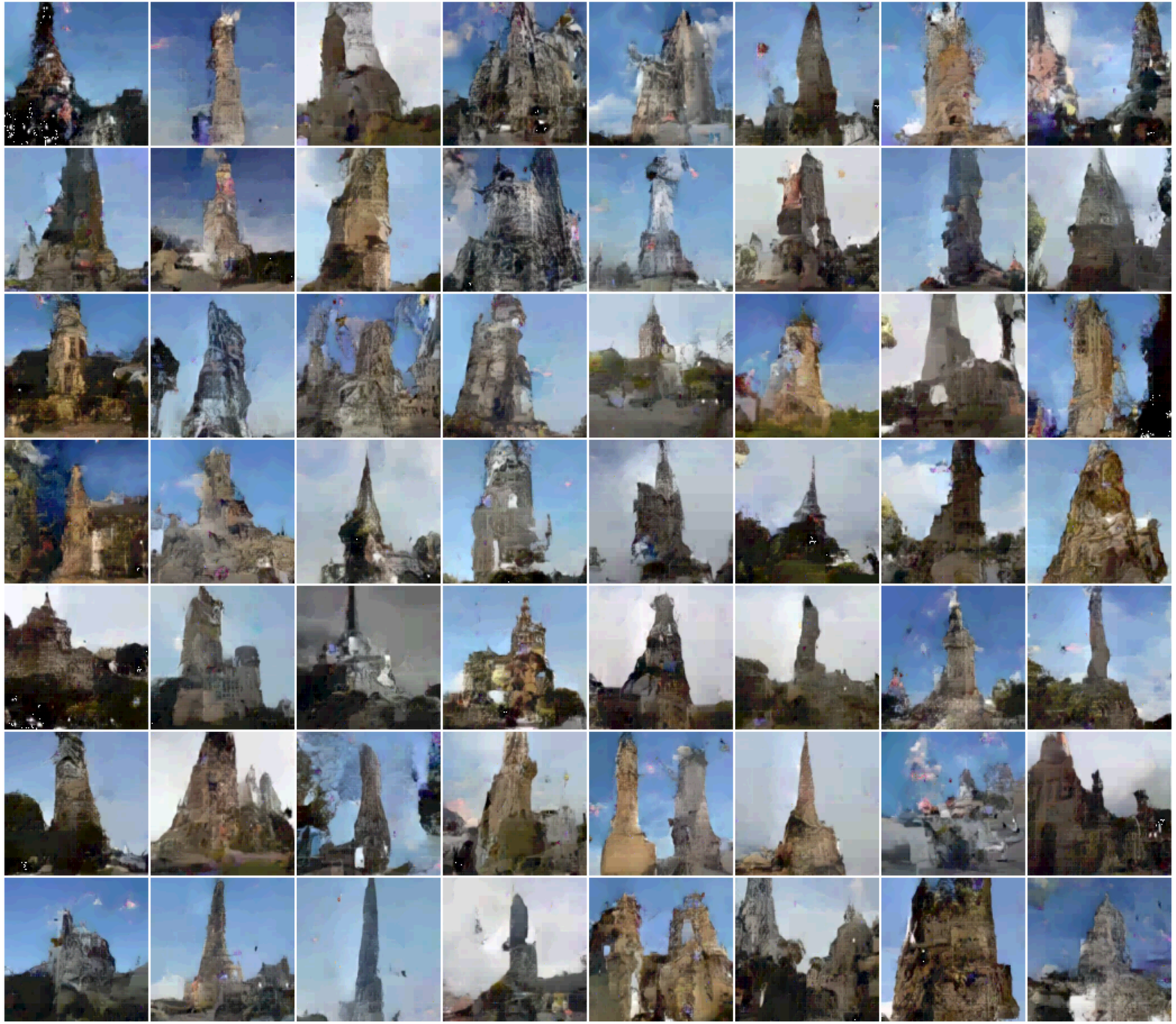}
    \caption{Random samples from 5-bit LSUN tower ($128 \times 128$).}
    \label{fig:lsuntower_suppmat}
    \vspace{-0.5em}
\end{figure*}

\begin{figure*}[t]
    \centering
   \includegraphics[width=\textwidth]{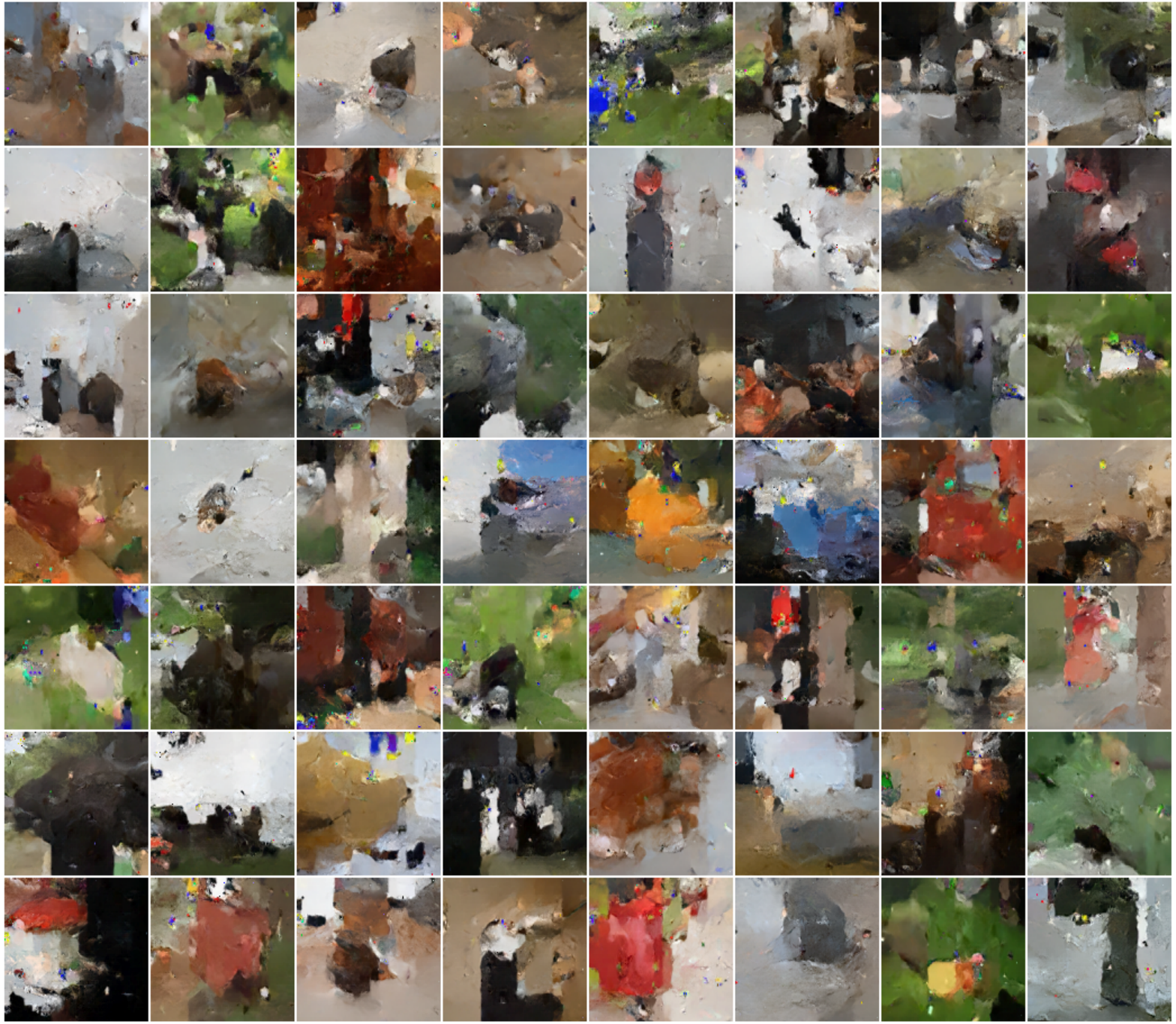}
    \caption{Random samples from 8-bit ImageNet ($128 \times 128$).}
    \label{fig:imagenet_suppmat}
    \vspace{-0.5em}
\end{figure*}

\begin{figure*}[t]
    \centering
   \includegraphics[width=0.75\textwidth]{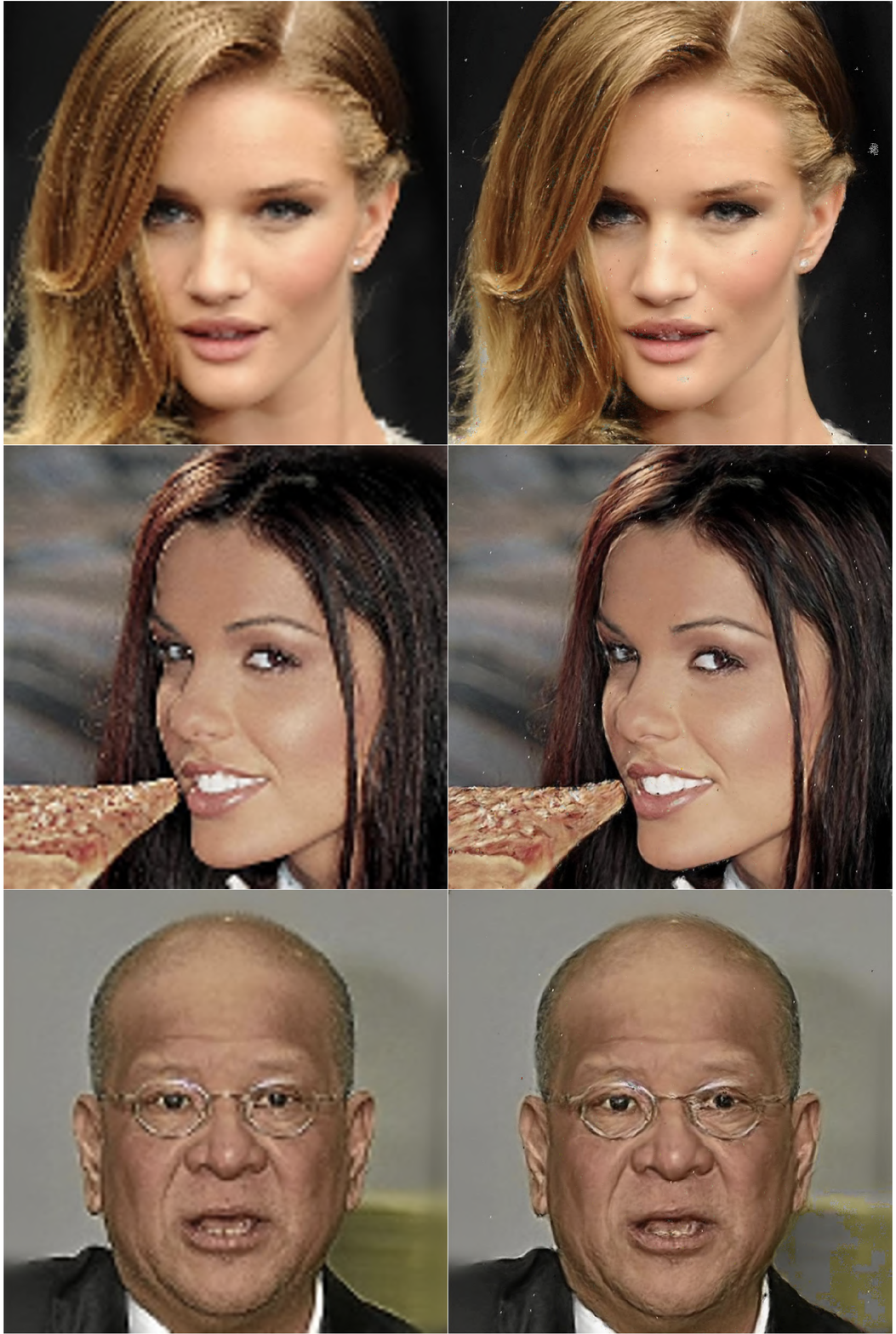}
    \caption{Super-resolution with PixelPyramids resizing a $128 \times 128$ image to $1024 \times 1024$ on the 8-bit CelebA-HQ ($1024 \times 1024$).}
    \label{fig:superres1_suppmat}
    \vspace{-0.5em}
\end{figure*}

\begin{figure*}[t]
    \centering
   \includegraphics[width=0.75\textwidth]{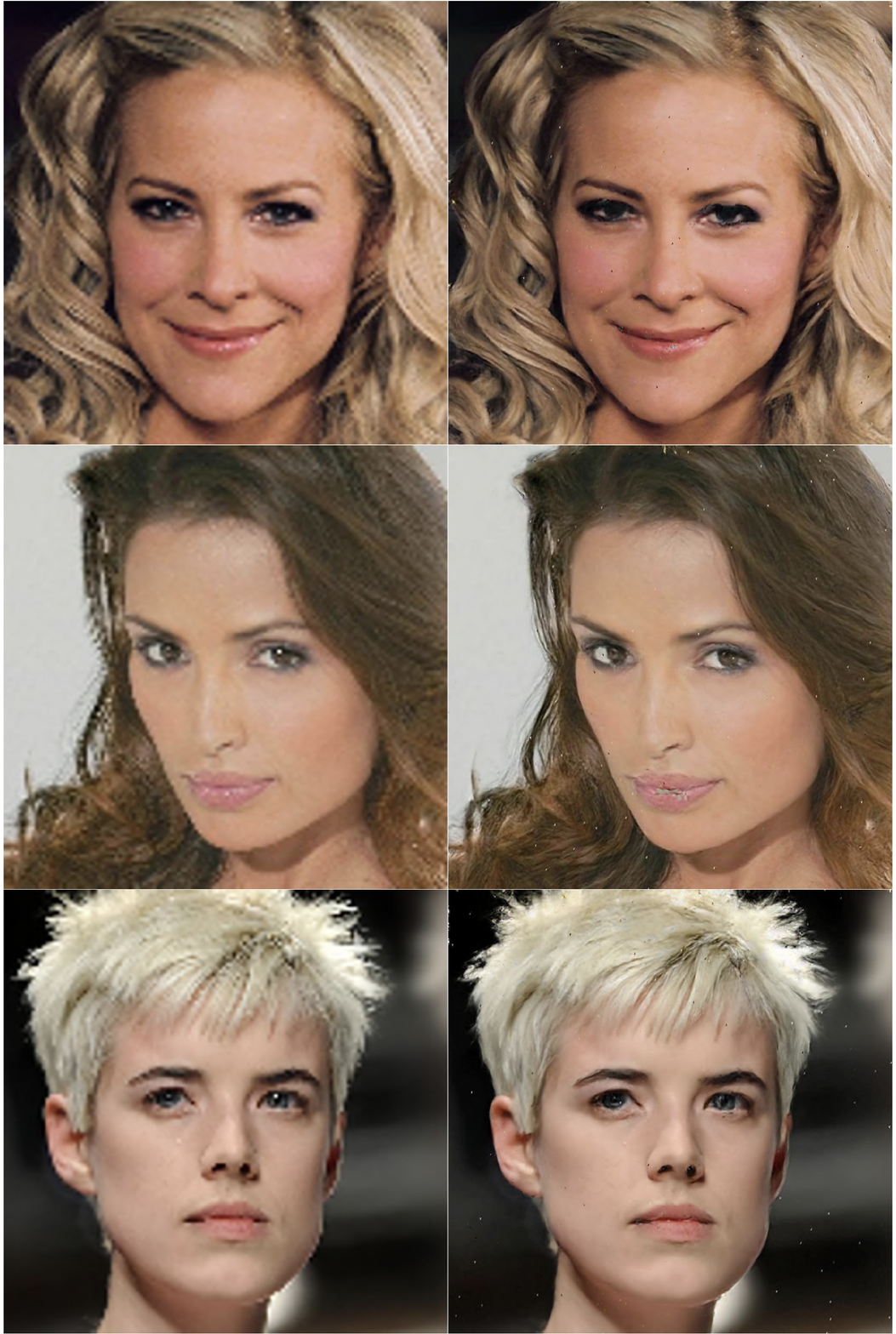}
    \caption{Super-resolution with PixelPyramids resizing a $128 \times 128$ image to $1024 \times 1024$ on the 8-bit CelebA-HQ ($1024 \times 1024$).}
    \label{fig:superres2_suppmat}
    \vspace{-0.5em}
\end{figure*}

We include additional qualitative examples obtained from our PixelPyramids on the high resolution datasets CelebA-HQ ($256 \times 256$; \cref{fig:celeba_suppmat}), CelebA-HQ ($1024 \times 1024$; \cref{fig:celeba1024_suppmat}), LSUN (bedroom, church outdoor, and tower, ($128 \times 128$); \cref{fig:lsunbedroom_suppmat,fig:lsunchurch_suppmat,fig:lsuntower_suppmat}), and ImageNet ($128 \times 128$; \cref{fig:imagenet_suppmat}).
PixelPyramids can synthesize high quality and diverse samples, capturing important visual properties in varied high-resolution datasets.
Furthermore, in \cref{fig:superres1_suppmat,fig:superres2_suppmat} we show the applicability of our PixelPyramids framework to the task of super-resolution, where fine details are iteratively added to the coarse input image ($128 \times 128$) at every level of PixelPyramids to generate a high-resolution output with a spatial resolution of  $1024 \times 1024$. We obtain an average PSNR($\uparrow$) of 27.25dB compared to 23.18dB for the baseline bicubic kernel \cite{LugmayrDGT20} and LPIPS($\downarrow$) of 0.28 compared to 0.51 with a bicubic kernel.


\end{document}